\documentclass{article}

\usepackage{arxiv}

\usepackage[utf8]{inputenc} % allow utf-8 input
\usepackage[T1]{fontenc}    % use 8-bit T1 fonts
\usepackage{hyperref}       % hyperlinks
\usepackage{url}            % simple URL typesetting
\usepackage{booktabs}       % professional-quality tables
\usepackage{amsfonts}       % blackboard math symbols
\usepackage{nicefrac}       % compact symbols for 1/2, etc.
\usepackage{microtype}      % microtypography
\usepackage{amsmath, amssymb}
\usepackage{subcaption}
\usepackage{cleveref}       % smart cross-referencing
\usepackage{lipsum}         % Can be removed after putting your text content
\usepackage{graphicx}
\usepackage{natbib}
\usepackage{doi}
\usepackage{multirow}
\usepackage{algorithm}
\usepackage{algorithmic}

\title{PCC: Pivotal Channel Compensation for Layer Pruning of Large Language Models}

\newif\ifuniqueAffiliation
\uniqueAffiliationtrue
\ifuniqueAffiliation
\author{
Xiangqian Wang \And
Yucheng Zhang \And
Zhaohong Jia \And
Kun Jing$^*$ \And
Chunchun Li$^*$
}
\fi

\hypersetup{
pdftitle={A template for the arxiv style},
pdfsubject={q-bio.NC, q-bio.QM},
pdfauthor={David S.~Hippocampus, Elias D.~Striatum},
pdfkeywords={First keyword, Second keyword, More},
}

\begin{document}
\maketitle

\begin{center}
School of Internet, Anhui University\\
No. 3 Feixi Road, Shushan District, Hefei, Anhui 230000, China\\[4pt]
\small $^*$ Corresponding author. \texttt{jingkun@ahu.edu.cn, licc@ahu.edu.cn}.\\
\end{center}

\begin{abstract}
	Treating independent layers as the minimal pruning unit allows layer pruning to efficiently eliminate many redundant parameters and achieve high compression rates. Layer pruning has become a promising and prevalent structured pruning technique for large language models (LLMs). However, multi-layer removal can lead to significant performance degradation for LLMs. Although some works have explored various compensation methods for pruned layers, they are inefficient and costly. Observing that the expression of a large proportion of information is concentrated on a small number of pivotal channels, we propose a novel post-pruning compensation method for layer pruning, called pivotal channel compensation (PCC). PCC can achieve significant performance improvement by compensating for a small number of channels through learnable linear transformations at the pruning interface. Besides, we propose a new evaluation metric, called channel-wise average performance improvement (CAPI), which can measure the efficiency of channel-wise compensation. Our experimental results across various LLMs demonstrate that PCC can achieve significant and consistent improvements and high CAPI on language modeling and zero-shot question answering for the state-of-the-art cross-layer and consecutive-layer pruning methods. Compared with existing compensation methods, PCC can achieve competitive results with fewer additional parameters and less training cost.
\end{abstract}

% keywords can be removed

\section{Introduction}
Large language models (LLMs) have drawn widespread attention in recent years due to their exceptional performance. However, their large size hinders deployment on edge devices, thereby limiting their practical applicability.
For the deployment of LLMs on the resource-constrained devices \cite{DBLP:journals/corr/abs-2402-06196,DBLP:journals/corr/abs-2401-00625,DBLP:journals/tacl/ZhuLLMW24}, researchers have devoted significant efforts to LLM compression techniques \cite{DBLP:conf/acl/TaoHZSJLLW22,DBLP:journals/corr/abs-2312-05821,DBLP:journals/nn/DongC24,DBLP:journals/nn/WangHGHGL26,DBLP:journals/nn/GuoHSZLTZCPM25}. Among these techniques, pruning methods \cite{DBLP:conf/nips/MaFW23,DBLP:conf/icml/FrantarA23,DBLP:conf/iclr/Sun0BK24,DBLP:journals/nn/TaoCGWWNZ25} have remained a research focus. The pruning methods can be categorized into structured pruning \cite{DBLP:conf/acl/Zhang0SYOYZ24,DBLP:conf/nips/GaoLHTSJH24,DBLP:journals/corr/abs-2302-03773,DBLP:conf/iclr/AshkboosCNHH24,DBLP:journals/nn/ZhangLWWW24} and unstructured pruning \cite{DBLP:conf/icassp/ShaoLQ24,DBLP:conf/emnlp/ZhaoSSYKWL24,DBLP:journals/corr/abs-2405-03594}. Due to the direct deletion of a large number of parameters without the need for specialized hardware support, structured pruning emerges as a more promising pruning paradigm. Compared with block- or module-level pruning methods, layer-level pruning methods efficiently offer more aggressive redundancy elimination in LLMs, driving the recent surge of interest in layer pruning.

Some layer pruning works \cite{DBLP:conf/icml/SongOKKKK24,DBLP:conf/acl/MenXZYWL0HC25} have proved that directly pruning redundant layers of LLMs is a feasible and efficient pruning method. They propose to use cosine distance or perplexity (PPL) to identify redundant layers. However, layer pruning methods that remove multiple intermediate layers simultaneously may result in substantial performance degradation \cite{DBLP:conf/iclr/ChenHZWL025}.
The performance degradation after layer pruning is primarily caused by the semantic gap resulting from concatenating the original discontinuous layers. Recently, some works \cite{DBLP:conf/iclr/ChenHZWL025,DBLP:conf/emnlp/YangC024,DBLP:conf/nips/ChenBYLZYHGHY25} propose to reconstruct the pruned layers with a small number of parameters to compensate for the performance degradation after layer pruning. However, existing compensation methods face challenges of time-consuming training, substantial parameter increase, and lack of interpretability.

\begin{figure*}[t]
\centering
\includegraphics[width=0.9\textwidth]{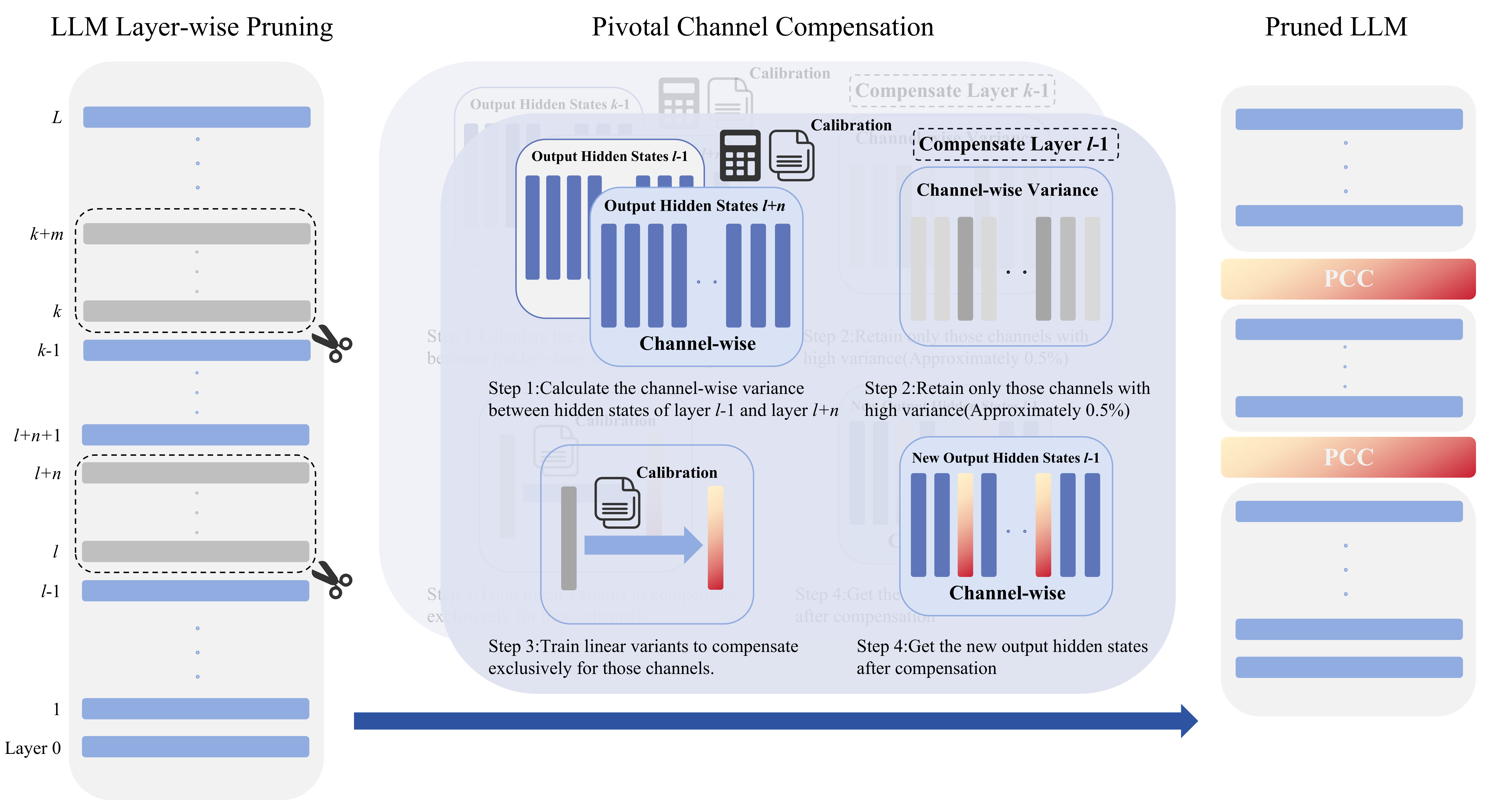} % Reduce the figure size so that it is slightly narrower than the column.
\caption{The overview of our proposed PCC.}
\label{fig1}
\end{figure*}

Motivated by the observation that a large proportion of information expression is concentrated in a small number of channels, we propose a novel pivotal channel compensation (PCC) method for layer pruning of LLMs, as shown in Figure~\ref{fig1}. Specifically, we find that merely 0.5\% of channels exhibit inter-layer variation vastly exceeding that of all others, and these pivotal channels play a decisive role in model performance. We propose to compensate for these pivotal channels at the pruning interface by learnable linear transformations trained with 1K calibration samples. This training process completes in just a few minutes on a single RTX 4090 GPU. Our experimental results demonstrate that for the state-of-the-art (SOTA) cross-layer pruning methods, our PCC method achieves an average PPL improvement of 3.58\% across LLaMA3-8B, LLaMA2-7B, LLaMA2-13B, and Baichuan2-7B models, by compensating for merely 0.5\% of the channels. For the SOTA consecutive-layer pruning methods across these models, PCC can achieve an average PPL improvement of 40.99\%. Compared with the SOTA full-channel compensation methods for layer pruning, PCC achieves competitive performance with extremely low cost. The PCC method also outperforms the SOTA compensation methods in terms of our proposed channel-wise average performance improvement (CAPI) metric. PCC also exhibits competitive results in zero-shot tasks. Overall, PCC can achieve substantial performance gains with few additional parameters and little cost.

Our contributions are as follows.
\begin{itemize}
    \item We reveal that inter-layer output variations in LLMs are highly concentrated in merely $\sim$0.5\% of channels, which exert substantial impacts on model performance and exhibit distinct layer-wise distributions. Furthermore, we observe that indiscriminate channel compensation can be detrimental, i.e., not all channels improve performance when compensated, with some even degrading language modeling capability. The phenomenon is overlooked by prior full-channel compensation approaches.
    \item To address this, we propose PCC, a novel layer pruning compensation method that selectively compensates only 0.5\% of channels with few extra parameters. PCC significantly outperforms inefficient full-channel compensation methods, enables efficient compensation in challenging cross-layer pruning scenarios where existing methods fail, and serves as a plug-and-play module compatible with existing layer pruning techniques.
    \item Furthermore, we introduce a novel evaluation metric that can quantify the contribution of compensated channels to the improvement of model performance, called CAPI. Extensive experiments demonstrate that PCC achieves superior CAPI scores across diverse LLMs compared to the SOTA methods.
\end{itemize}

\section{Related Work}
Many works propose various metrics for layer pruning. Given that language modeling is the fundamental task, SLEB \cite{DBLP:conf/icml/SongOKKKK24} proposes to use PPL to prune redundant layers. LaCo \cite{DBLP:conf/emnlp/YangC024} identifies consecutive layers to be merged into their predecessors based on the cosine distance of output hidden states before and after compression. ShortGPT \cite{DBLP:conf/acl/MenXZYWL0HC25} identifies and removes redundant layers based on the BI score, a layer importance evaluation metric based on cosine distance. The sliding layer merging method \cite{DBLP:journals/corr/abs-2502-19159} proves the existence of redundant layers in LLMs using Hilbert space, and then dynamically removes redundant layers through the sliding windows. Although these methods can achieve efficient pruning of redundant layers and higher pruning ratios, the performance degrades rapidly as the pruning rate increases. Some works further propose performance compensation for layer pruning methods. LaCo \cite{DBLP:conf/emnlp/YangC024} incorporates the parameter differences between subsequent layers and the preceding layer into the latter to update its weights, and substitutes these consecutive layers with the updated single layer. LLM-Streamline \cite{DBLP:conf/iclr/ChenHZWL025} employs 30,000 calibration samples to train Transformer and FFN layers separately for replacing the removed layers. LINEARPATCH \cite{DBLP:conf/nips/ChenBYLZYHGHY25} first utilizes Hadamard transformation to constrict outliers in particular tokens and then takes channel-wise scaling at the pruning interface to align the activation magnitudes.

Our PCC method further explores efficient compensation methods for pruned layers within the layer pruning framework. PCC addresses the key limitations of existing post-pruning compensation methods, including high training costs, excessive additional parameters, lack of interpretability, and difficulties in cross-layer compensation. Compared to the sliding layer merging and Laco methods, PCC offers a more rigorous theoretical foundation for post-pruning compensation and superior performance. Compared with these existing works, our method achieves effective performance compensation with fewer additional parameters and lower training cost.

\section{Methodology}
\subsection{Preliminary of Layer Pruning for LLMs}
The previous works \cite{DBLP:journals/corr/abs-2411-15558,DBLP:conf/acl/MenXZYWL0HC25} have shown that there exist redundant layers that have weak information representation abilities in LLMs. Pruning redundant layers results in less performance degradation than removing other layers. As layer pruning methods mature, cosine distance and PPL have become two widely used metrics for evaluating layer importance
\begin{equation}
\text{Metric}_{Cos} = 1 - \frac{H_l^\text{In} \cdot H_l^\text{Out}}{\|H_l^\text{In}\| \|H_l^\text{Out}\|}
\label{equ1}
\end{equation}
\begin{equation}
\text{Metric}_{PPL} =\exp(-\mathbb{E}\left[\log p_{M_l}(w_k \mid w_{<k})\right]-\exp(-\mathbb{E}\left[\log p_M(w_k \mid w_{<k})\right])
\label{equ2})
\end{equation}
where $H_l^\text{In}$ and $H_l^\text{Out}$ denote the input and output hidden states of the $l$-th layer respectively; $M_l$ represents the targeted model with the $l$-th layer pruned; $p_{M}(w_k \mid w_{<k})$ denotes the probability of next token $w_k$ for the model $M$ given the prefix tokens $w_{<k}$.

As shown in Table~\ref{table1}, we provide an overview of redundant layers on the mainstream LLMs identified by different layer pruning methods with cosine distance and PPL metrics. The methods with cosine distance tend to prune consecutive layers, while the methods with PPL exhibit a preference for cross-layer pruning. The time-consuming training and substantial parameter increase render existing post-pruning compensation methods inefficient in the cross-layer pruning scenario.

\begin{table*}[ht]
    \centering
    \begin{tabular}{c c c} % 去除竖线，改为四列左对齐
        \toprule % 顶部粗线
        \textbf{Model} & \textbf{Method} & \textbf{Pruned Layers}  \\
        \midrule % 中间线
        \multirow{2}{*}{\centering LLaMA2-7B} & SLEB (PPL) & 10,11,14,15,23,24,27 \\
        & LLM-Streamline (Cos) & 23,24,25,26,27,28,29 \\
        \midrule % 中间线
        \multirow{2}{*}{\centering LLaMA2-13B} & SLEB (PPL) & 2,12,13,14,27,29,31,33 \\
        & LLM-Streamline (Cos) & 26,27,28,29,30,31,32,33,34,35 \\
        \midrule % 中间线
        \multirow{2}{*}{\centering LLaMA3-8B} & SLEB (PPL) & 9,10,11,12,23,25,26 \\
        & LLM-Streamline (Cos) & 23,24,25,26,27,28,29 \\
        \midrule % 中间线
        \multirow{2}{*}{\centering Baichuan2-7B} & SLEB (PPL) & 
        4,13,14,15,16,26,27\\
        & LLM-Streamline (Cos) & 
        23,24,25,26,27,28,29\\
        \bottomrule % 底部粗线
\end{tabular}
\caption{The pruned layer indices based on PPL and cosine distance (Cos).}
\label{table1}
\end{table*}

\paragraph{Discussion 1: Dilemmas of Existing Layer Pruning Metrics}
Although a small cosine distance can indicate strong correlations in semantics, syntax, and context, this does not equate to semantic equivalence. Cosine distance can only reflect the directional similarity between the hidden representation vectors but ignores their differences in magnitude. Therefore, cosine distance may inaccurately overestimate the real importance of the layers, and its standalone use remains methodologically insufficient.
Some researchers have adopted PPL as their evaluation metric. A lower PPL indicates a closer alignment between the predicted and real distributions on the fixed calibration set. However, this is highly dependent on the choice of calibration sets. The same layer exhibits varying expressive capacities when facing different input, leading to discrepancies in identifying redundant layers using the PPL metric with different calibration data.
In short, these two mainstream metrics still have inherent limitations that cannot be avoided. Besides, regardless of which layers are pruned, performance degradation is inevitable.
Therefore, we argue that the post-pruning performance compensation is a more promising approach for enhancing layer pruning, and pivot from identifying redundant layers toward post-pruning performance compensation. 

\begin{table*}[ht]
\setlength{\tabcolsep}{1mm}
    \centering
    \begin{tabular}{c c c c c} % 去除竖线，改为四列左对齐
        \toprule % 顶部粗线
        Model & Pruned Layers & $V_{\text{mean}}$ & $V_{\text{top 0.5\%}}$ & Ratio \\
        \midrule % 中间线
        \multirow{5}{*}{\centering LLaMA2-7B} & 10,11 & 14.22 & 82.26 & 5.78$\times$ \\
        & 14,15 & 30.48 & 227.73 & 7.47$\times$ \\
        & 23,24 & 117.85 & 2229.21 & 18.91$\times$ \\
        & 27 & 76.86 & 315.17 & 4.10$\times$ \\
        & 23-29 & 839.08 & 31053.06 & 37.01$\times$ \\
        \midrule % 中间线
        \multirow{7}{*}{\centering LLaMA2-13B} & 2 & 0.79 & 19.45 & 24.62$\times$ \\
        & 12,13,14 & 38.25 & 302.28 & 7.90$\times$ \\
        & 27 & 53.95 & 354.79 & 6.58$\times$ \\
        & 29 & 53.94 & 310.16 & 5.75$\times$ \\
        & 31 & 60.43 & 425.41 & 7.04$\times$ \\
        & 33 & 71.86 & 826.61 & 11.50$\times$ \\
        & 26-35 & 986.18 & 41660.22 & 42.24$\times$ \\
        \midrule % 中间线
        \multirow{5}{*}{\centering LLaMA3-8B} & 9-12 & 3.04 & 28.58 & 9.40$\times$ \\
        & 23 & 4.81 & 89.35 & 18.58$\times$ \\
        & 25,26 & 13.56 & 377.72 & 27.86$\times$ \\
        & 22-28 & 73.06 & 3722.04 & 50.95$\times$ \\
        & 23-29 & 83.40 & 2931.03 & 35.14$\times$ \\
        \midrule % 中间线
        \multirow{4}{*}{\centering Baichuan2-7B} & 4 & 0.64 & 7.00 & 10.94$\times$ \\
        & 13-16 & 36.78 & 473.53 & 12.87$\times$ \\
        & 26,27 & 190.25 & 5508.88 & 28.96$\times$ \\
        & 23-29 & 1166.86 & 69877.86 & 59.89$\times$ \\
        \bottomrule % 底部粗线
\end{tabular}
\caption{The inter-layer variance values of all channel values ($V_{\text{mean}}$) and top 0.5\% of channel values by variance ($V_{\text{top 0.5}}$) in different LLMs.}
\label{table2}
\end{table*}

\subsection{Pivotal Channels}
The key performance degradation \cite{DBLP:conf/nips/ChenBYLZYHGHY25,DBLP:conf/iclr/ChenHZWL025} after layer pruning is that directly concatenating non-consecutive layers may disrupt the alignment of the original feature space, resulting in non-consecutive propagation of semantic dependencies at the pruning interface. Layer replacement, merging, and finetuning are the mainstream compensation methods for the difference at the pruning layer. However, these methods perform full-channel compensation at the pruning interfaces, which are computationally expensive to train and incur substantial additional parameters. For training-free compensation methods, as they rely on heuristic prior knowledge and strict assumptions, they often yield suboptimal performance.

 We observe that the inter-layer variation in a small subset of channel values is significantly higher than that of other channels. Different from the discovery of massive activations \cite{DBLP:journals/corr/abs-2402-17762}, we do not care about the activation magnitude of individual channel outputs but the significance of channel variations. We further extend the observations of massive activations, our analysis expanding the number of pivotal channels from 3 $\sim$ 4 to approximately 0.5\% of all channels. We randomly select a set of sentences as inputs to the mainstream LLMs and calculate the channel-wise variance between different layers. As shown in Table~\ref{table2}, we observe that, regardless of which layers are pruned, approximately 0.5\% of total channels exhibit higher variance than other channels.

 \begin{figure*}[ht]
    \centering
    % 第一行子图
    \begin{subfigure}{0.45\linewidth}
        \centering
        \includegraphics[width=\linewidth]{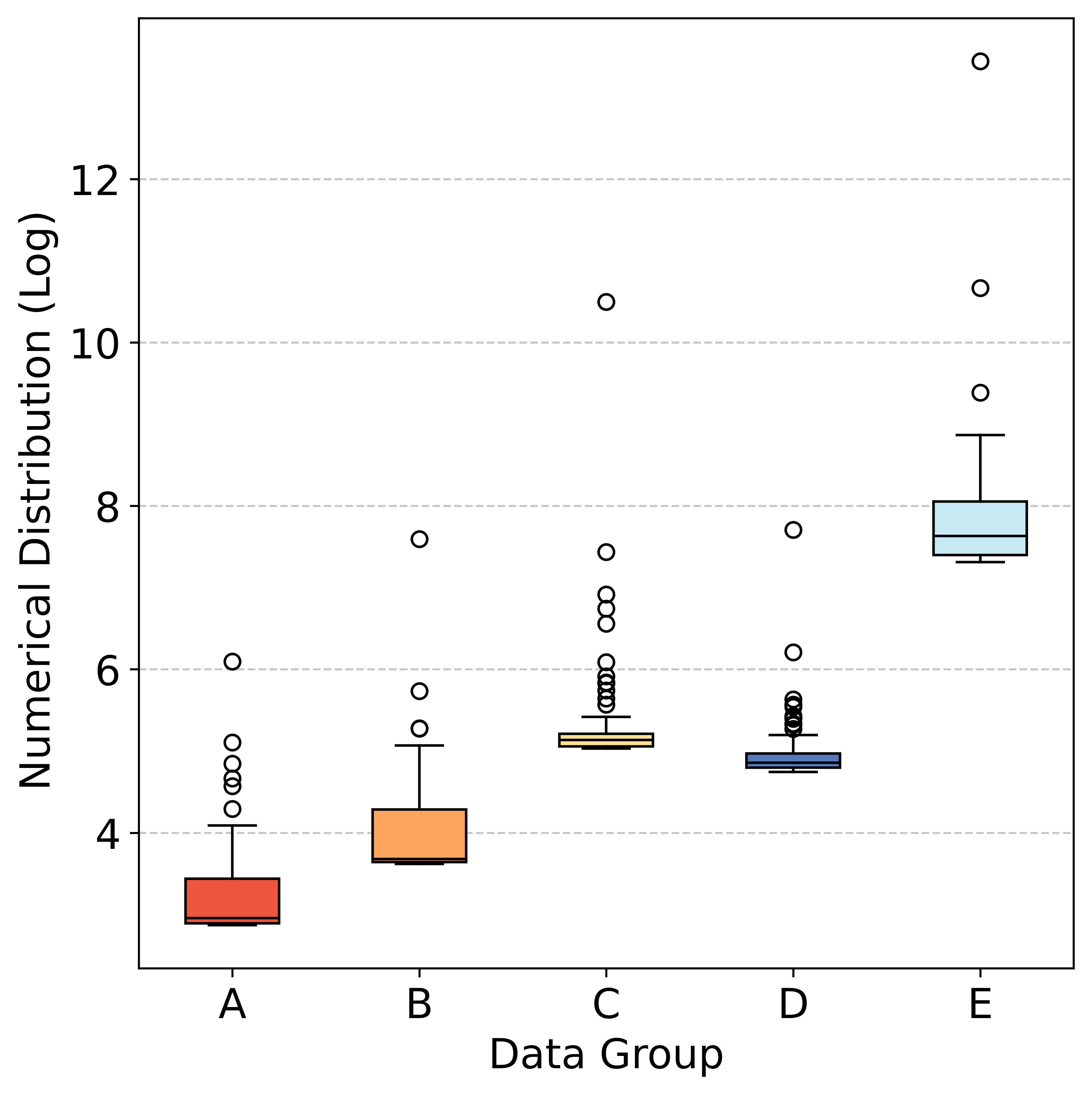}
        \caption{The distribution of channel variance values across different groups of layers.}
        \label{fig:sub1}
    \end{subfigure}
    \hfill
    \begin{subfigure}{0.45\linewidth}
        \centering
        \includegraphics[width=\linewidth]{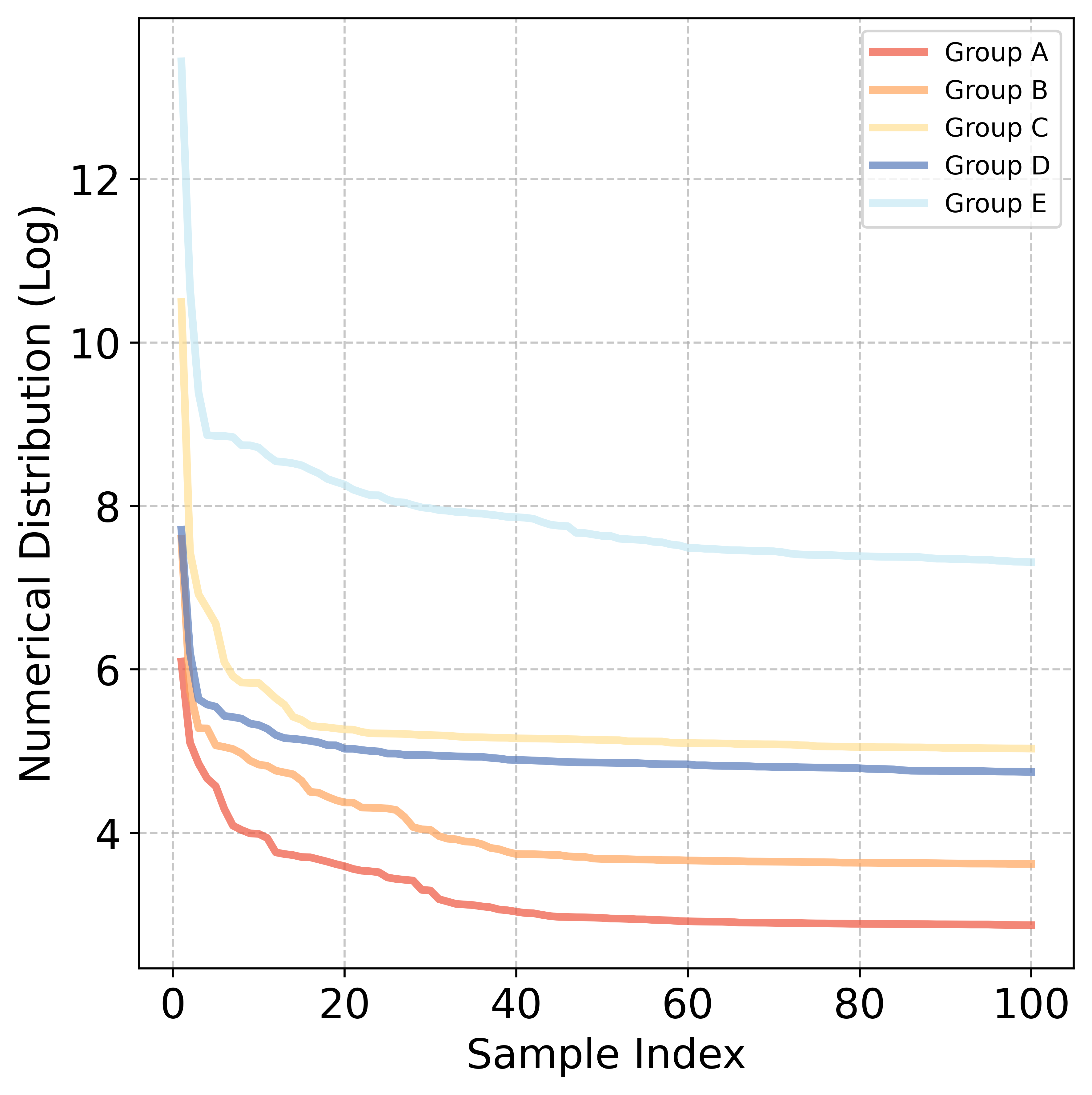}
        \caption{The variation trend of inter-layer variance values across different groups in descending order of magnitude.}
        \label{fig:sub2}
    \end{subfigure}
    \caption{The variance values of inter-layer channels on LLaMA2-7B. We report the top 100 values and all variance values are log-transformed.
    Groups A to D represent intermediate layers pruned via SLEB method, while Group E denotes those identified by the LLM-Streamline method. For any given data group, its top 20 variance values (accounting for 0.5\% of the total channels) are consistently significantly higher than the others.}
    \label{fig2}
    \end{figure*}

\begin{figure*}[t]
    \centering
    % 第一行子图
    \begin{subfigure}{0.45\linewidth}
        \centering
        \includegraphics[width=\linewidth]{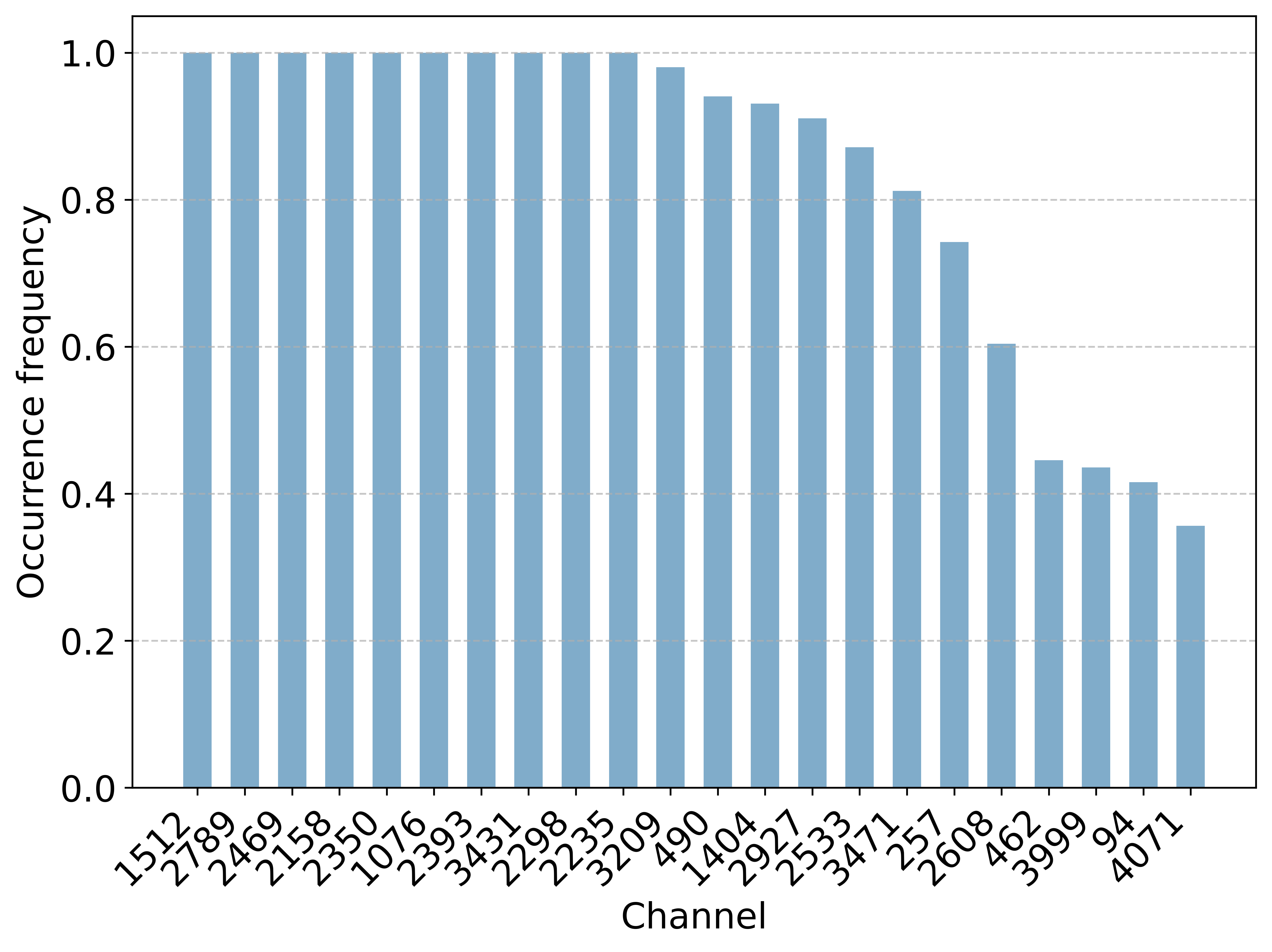}
        \caption{The occurrence frequency of pivotal channels between layer9 and layer11.}
        \label{fig:sub3}
    \end{subfigure}
    \hfill
    \begin{subfigure}{0.45\linewidth}
        \centering
        \includegraphics[width=\linewidth]{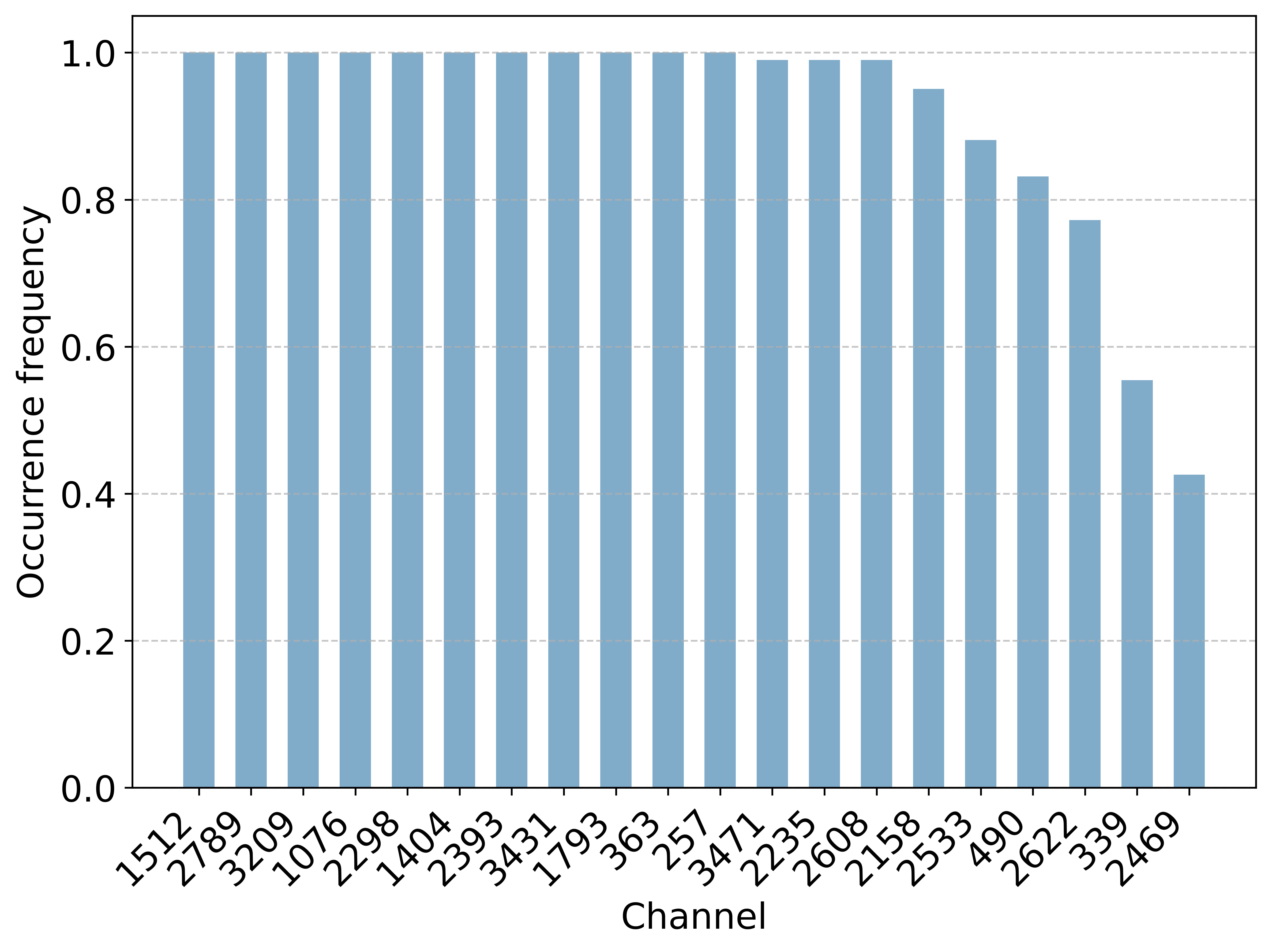}
        \caption{The occurrence frequency of pivotal channels between layer13 and layer15.}
        \label{fig:sub4}
    \end{subfigure}
    \smallskip % 中等大小的垂直间距（比\vspace更符合LaTeX排版习惯）
    % 第二行子图
    \begin{subfigure}{0.45\linewidth}
        \centering
        \includegraphics[width=\linewidth]{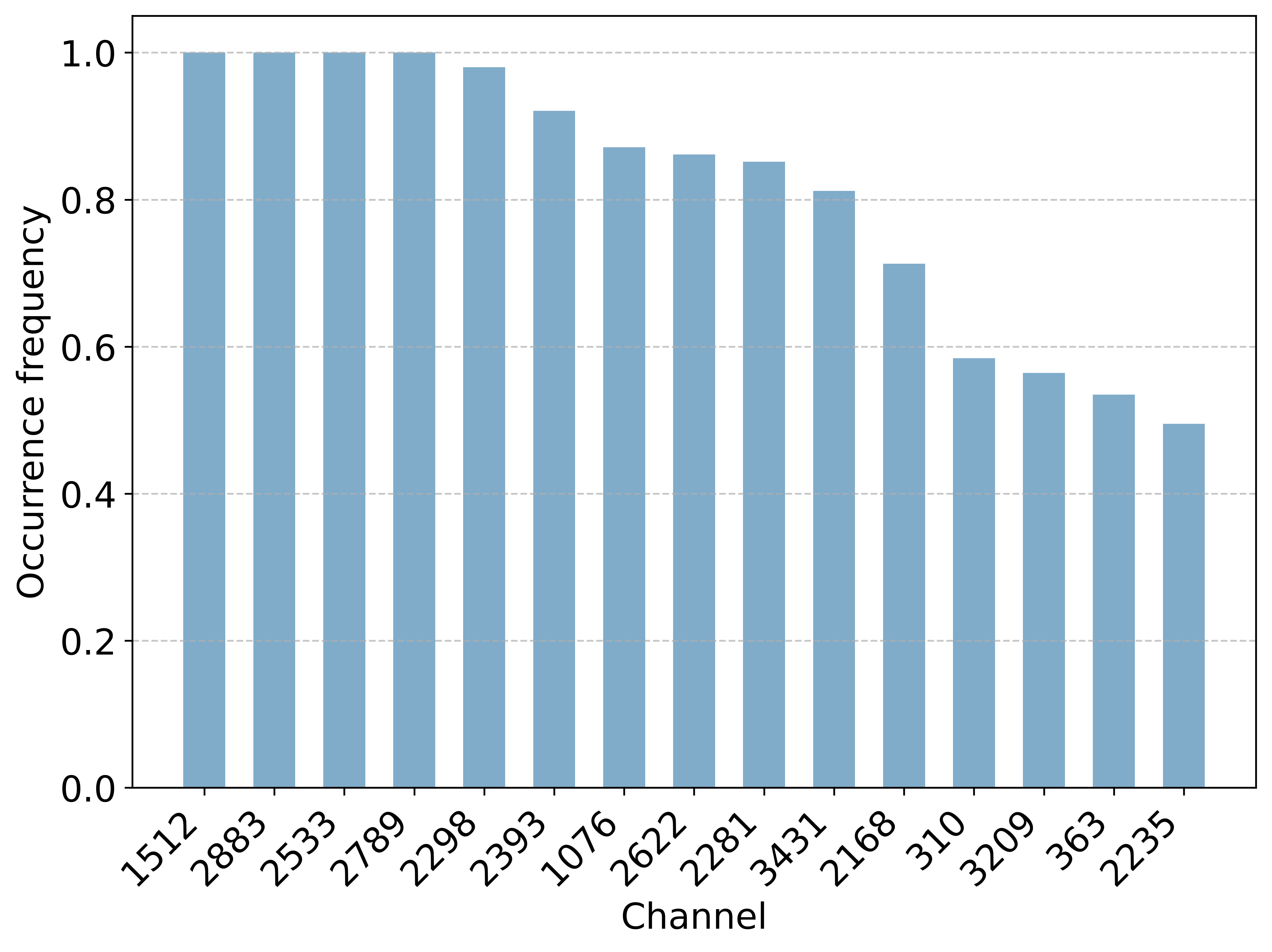}
        \caption{The occurrence frequency of pivotal channels between layer22 and layer24.}
        \label{fig:sub5}
    \end{subfigure}
    \hfill
    \begin{subfigure}{0.45\linewidth}
        \centering
        \includegraphics[width=\linewidth]{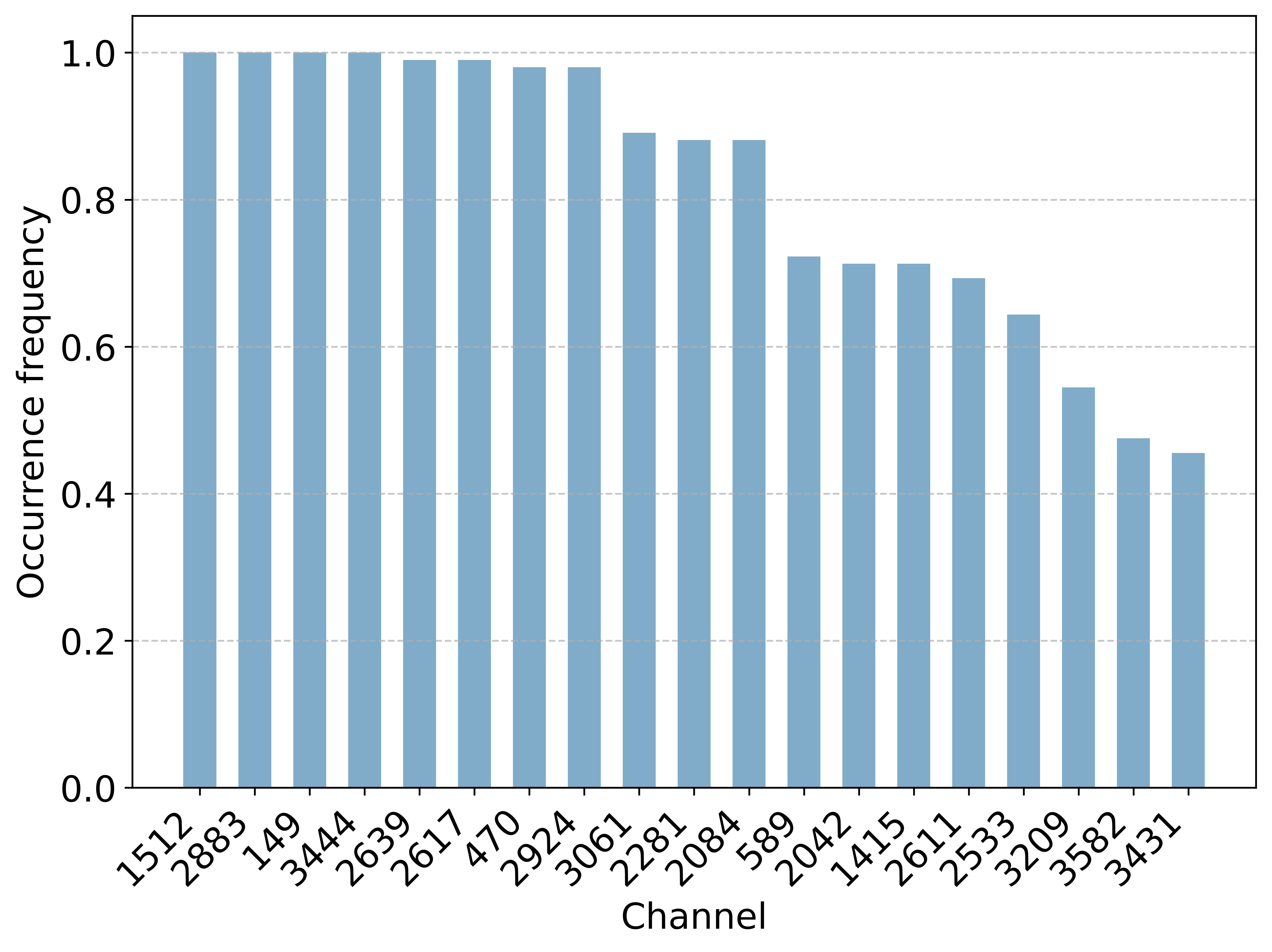}
        \caption{The occurrence frequency of pivotal channels between layer26 and layer27.}
        \label{fig:sub6}
    \end{subfigure}
    \caption{The occurrence frequency of different channels as pivotal channels at different pruning interfaces. We randomly select 100 sentences as inputs.}
    \label{fig3}
  \end{figure*}

Taking the LLaMA2-7B model as an example, the output hidden states of each layer contain a total of 4096 channels. 
In Figure~\ref{fig2}, it can be clearly observed that the variance values of these pivotal channels are significantly higher than those of other channels. To further validate scalability, we analyze 100 random input sequences and report the indices of the top 0.5\% of channels by variance in output hidden states across layers. As shown in Figure~\ref{fig3}, statistical analysis of their occurrence frequencies reveals a universal observation: for different input sequences, the output hidden states exhibit high variance on pivotal channels. The detailed discussion on other LLMs can be found in ~\ref{app1}.
\subsection{Pivotal Channel Compensation}
\paragraph{Discussion 2: Dilemmas of Existing Compensation Methods} There are two mainstream methods for performance compensation after layer pruning: fine-tuning and layer compensation. For fine-tuning, though parameter-efficient fine-tuning (PEFT) like LoRA can accelerate the whole process, frequent fine-tuning remains impractical for LLMs with a minimum scale of one billion parameters. The second approach retains the parameters of remaining layers intact and retrains an entire layer using the input and output hidden states of the pruned layers as training data to compensate for performance drops. LLM-Streamline achieves SOTA performance under this paradigm; however, even when training a single layer, the computational and temporal overheads of this process remain prohibitive. Furthermore, the introduction of new parameters diminishes the original pruning ratio. More critically, due to the need for substantial additional parameters, layer substitution compensation methods are limited to consecutive layer pruning scenarios, and struggle to handle cross-layer pruning.

The above analysis demonstrates that there is a distinct observation in the pivotal channels: the channels that exhibit high variance at the pruning interface usually remain consistent. The channels with high variance indicate that they carry more abundant information and the corresponding channels at the pruning interfaces have stronger expressive capacity. Therefore, it is justified to prioritize the compensation of such channels with strong expressive capabilities.
\begin{table*}[ht]
    \setlength{\tabcolsep}{1mm}
    \centering
    \begin{tabular}{c c c c c c c c c c c} % 去除竖线，改为四列左对齐
    \toprule % 顶部粗线
        Dimension & 1512 & 2883 & 149 & 3444 & 2639 & 2617 & 470 & 2924 & 3061 & 2281  \\
        \midrule % 中间线
        Pearson & 0.993 & 0.994 & 0.965 & 0.970 & 0.964 & 0.958 & 0.971 & 0.965 & 0.965 & 0.985   \\
        \bottomrule % 底部粗线
\end{tabular}
\caption{The Pearson correlation coefficients of pivotal channels.}
\label{table3}
\end{table*}

Motivated by the residual connection inherent that can amplify linear correlations between features in Transformer architecture, a simple compensation method is to train a linear transformation for each pivotal channel. To verify the linear correlation of inter-layer channels, we compute their Pearson correlation coefficients. The Pearson correlation coefficient serves as an important indicator for quantifying the linear correlation between two vectors. To be specific, when the Pearson correlation coefficient of two vectors exceeds 0.8, they are considered to be strongly linearly correlated. Consequently, if the analysis of the Pearson correlation coefficient reveals that the pivotal channels of these output hidden states have high linear correlation, then linear transformations could be employed in lieu of complex neural networks to perform targeted compensation for these pivotal channels. Table~\ref{table3} reports the mean Pearson correlation coefficients of several pivotal channels across different input instances. Our analysis on other layers and LLMs can be found in ~\ref{app2}, and these results consistently indicate strong linear correlations between the pivotal channels.

Based on this conclusion, we introduce our core compensation mechanism within the PCC framework: linear compensations are applied exclusively to these pivotal channels to restore their output before pruning. Since linear compensation can only serve as an approximation of the behavior of the original layer, inaccuracy is inevitable, especially for channels with insignificant variations. Therefore, we argue that current compensation methods that perform linear compensation on all channels are suboptimal. 

We observe that compensating for a subset of output channels fails to effectively enhance the model’s language modeling capability; on the contrary, it may even degrade the model performance. This aspect has been underexplored in previous full-channel compensation works. So when we get the linear compensation coefficients, we test the linear compensation effects on validation sets and discard some of them with negative effects. We provide a detailed description in the ablation study and show that linear compensation retains more valid compensation channels compared to scaling compensation. The PCC algorithm is summarized in Algorithm~\ref{alg1}. As shown in Lines 1-4 of Algorithm 1, we first extract the output hidden states from the l-th layer and the (l+n-1)-th layer of the model. For each token position and channel, we calculate the variance between the outputs of the two layers, then sum the variances across all token positions to obtain the total variance of each channel. In Lines 5-8 of Algorithm 1, we sort the channel variances in descending order and obtain the sorted channel indices. We then select the top 0.5\% of all channels with the highest variances from the sorted sequence as the pivotal channels, and in Lines 9-11, we adopt 10,000 calibration samples to extract hidden states from the two layers corresponding to each pivotal channel. The TrainLinearCompensation function serves as a simple linear regression model to minimize the difference between the compensated outputs and the original features of the target layer.

\subsection{Channel-wise Average Performance Improvement}
Existing works evaluate pruned LLMs using PPL on language modeling and accuracy on zero-shot question-answering, alongside the number of additional parameters for compensation. However, these metrics merely capture final performance without illuminating the critical trade-off between compensation cost and performance gain. Ignoring the additional parameters and computational cost introduced by compensation methods, such evaluation inherently favors parameter-heavy methods, directly contradicting the fundamental goal of model compression. Against this backdrop, we proposed a metric to quantify the trade-off between gain and cost. The most straightforward approach to evaluate this trade-off is to use the ratio of performance gain to compensated parameters. Given that all our comparative experiments focused on channel compensation, we replaced the denominator with the number of compensated channels to more intuitively compare the compensation effect and contribution of each channel, and accordingly proposed our evaluation metric channel-wise average performance improvement (CAPI),
\begin{equation}
    CAPI = \frac{\text{Performance Improvement}}{\text{Number of Compensation Channels}}
\label{equ3}
\end{equation}

Since all methods in this work adopted linear transformation for channel compensation, the number of parameters for each compensated channel is nearly identical. This revision is therefore reasonable. Besides, as the ratio of performance gain to the number of compensated channels, CAPI offers clear guidance for selecting the optimal proportion of compensated channels. CAPI serves as a complementary metric for evaluating the efficiency of compensation methods for layer pruning. By quantifying performance improvement per channel compensation cost, it enables direct comparison of cost-effectiveness: larger values indicate superior efficiency in achieving performance gains.

\begin{algorithm}[t]
\caption{PCC}
\label{alg1}
\textbf{Input}:  Number of channels $C$, Output hidden states $H$, Pruning interface $l$ and Pruning layers $n$.\\
\textbf{Output}: Linear compensation coefficient: scaling $a_i$ and bias $b_i$. \\
\begin{algorithmic}[1] %[1] enables line numbers
\FOR{$i=0$ to $C$-1}
\STATE// Calculate per-channel variances
\STATE $variance_i$ = $\text{Variance} (H_{l}[:,i], H_{l+n}[:,i]);$    
\ENDFOR
\STATE// Sort the variance values of all channels
\STATE S = ArgSort (variance, descending=True);
\STATE// Select top 0.5\% of channels by variance
\STATE S = S[:0.5\%$\times$C];
\FOR{$i$ in S}
\STATE $a_i$, $b_i$ = TrainLinearCompenstaion($H_{l}[:,i]$)
\ENDFOR
\STATE Integrate lightweight linear transformations into the pruned model, applying channel-specific projections to pivotal channels during each forward inference.
\end{algorithmic}
\end{algorithm}

\section{Experiments}
\subsection{Setup and Evaluation}

To validate the effectiveness of our proposed PCC method, we conducted three main experiments. We evaluated the performance of different pruning methods with or without compensation across different LLMs on the WikiText2 dataset of langauge modeling and the 9 mainstream datasets of zero-shot question answering (QA).
These pruning methods include PPL-based SLEB and cosine distance-based methods: ShortGPT and LLM-Streamline (LLM-S).
We denote the pruning method of LLM-Streamline without compensation as LLM-S[None].
We also compared our proposed PCC with SliceGPT, a representative method in structured pruning, and LINEARPATCH, the SOTA full-channel compensation method.
Furthermore, to validate the plug-and-play compatibility of PCC, we evaluate the performance of different pruning methods compensated by LINEARPATCH and PCC across different LLMs on the WikiText2 dataset of langauge modeling.

All experiments are conducted across four prevailing open-source LLMs, including the Llama3-8B, Llama2-7B, Llama2-13B, and Baichuan2-7B. For the Llama2-7B, Llama3-8B, and Baichuan2-7B models, 7 layers are pruned from the original 32-layer architecture; for the Llama2-13B model, 10 layers are pruned from the original 40-layer architecture. They are strictly following the SLEB \cite{DBLP:conf/icml/SongOKKKK24} and the LLM-Streamline \cite{DBLP:conf/iclr/ChenHZWL025} methods. We train the learnable linear compensation modules using the Adam optimizer with a learning rate of 1e-3 for 10,000 steps.

We evaluated the performance of the models on language modeling and zero-shot question answering tasks. For language modeling, we use PPL to evaluate the model text generation performance on the WikiText2 dataset, and CAPI to quantify the trade-off between the compensation cost and performance improvement of models. For zero-shot question answering, we evaluate the accuracies on 9 mainstream datasets, including ARC-Challenge, ARC-Easy, BoolQ, HellaSwag, PIQA, WinoGrande, WSC, MNLI, and ANLI.

\subsection{Main Results}
\subsubsection{Results on Language Modeling}

\begin{table*}[h!]
    \centering
    \setlength{\tabcolsep}{1mm}
    \begin{tabular}{c c c c c c c} % 去除竖线，改为四列左对齐
        \toprule % 顶部粗线
        Model & $L_p$/$L_t$ & Ratio & Method & PPL & Imp.(\%) & CAPI \\
        \midrule % 中间线
        \multirow{9}{*}{\centering LLaMA3-8B} & 0/32 & - & Dense & 6.13 & - & - \\
        \cmidrule{2-7} % 中间线
        & 7/32 & 19.01 & SLEB & 13.11 & - & - \\
        & 7/32 & 19.01 & PCC & \textbf{12.42} & \textbf{5.26} & \textbf{0.1384} \\
        \cmidrule{2-7} % 中间线
        & 7/32 & 19.01 & ShortGPT & 57.77 & - & - \\
        & 7/32 & 18.80 & LINEARPATCH & \textbf{25.35} & \textbf{56.12} & 0.0137 \\
        & 7/32 & 19.01 & PCC & 27.74 & 51.98 & \textbf{4.3317} \\
        \cmidrule{2-7} % 中间线
        & 7/32 & 19.01 & LLM-S[None] & 2305.17 & - & - \\
        & 7/32 & 18.80 & LINEARPATCH & 72.12 & 96.87 & 0.0236 \\
        & 7/32 & 19.01 & PCC & \textbf{64.82} & \textbf{97.19} & \textbf{4.2256} \\
        \midrule % 中间线
        \multirow{6}{*}{\centering LLaMA2-7B} & 0/32 & - & Dense & 5.47 & - & - \\
        \cmidrule{2-7} % 中间线
        & 7/32 & 21.02 & SLEB & 9.14 & - & - \\
        & 7/32 & 21.02 & PCC & \textbf{8.91} & \textbf{2.52} & \textbf{0.0899} \\
        \cmidrule{2-7} % 中间线
        & 7/32 & 21.02 & LLM-S[None]/ShortGPT & 18.45 & - & - \\
        & 7/32 & 20.78 & LINEARPATCH & \textbf{13.27} & \textbf{28.07} & 0.0068 \\
        & 7/32 & 21.02 & PCC & 14.50 & 21.41 & \textbf{1.0195} \\
        \midrule % 中间线
        \multirow{6}{*}{\centering LLaMA2-13B} & 0/40 & - & Dense & 4.88 & - & - \\
        \cmidrule{2-7} % 中间线
        & 8/40 & 19.50 & SLEB & 8.17 & - & - \\
        & 8/40 & 19.50 & PCC & \textbf{7.90} & \textbf{3.30} & \textbf{0.0635} \\
        \cmidrule{2-7} % 中间线
        & 10/40 & 24.37 & LLM-S[None]/ShortGPT & 9.77 & - & - \\
        & 10/40 & 24.17 & LINEARPATCH & 8.74 & 10.54 & 0.0020 \\
        & 10/40 & 24.37 & PCC & \textbf{8.69} & \textbf{11.05} & \textbf{0.5023} \\
        \midrule % 中间线
        \multirow{6}{*}{\centering Baichuan2-7B} & 0/32 & - & Dense & 6.03 & - & - \\
        \cmidrule{2-7} % 中间线
        & 7/32 & 18.87 & SLEB & 10.78 & - & - \\
        & 7/32 & 18.87 & PCC & \textbf{10.43} & \textbf{3.25} & \textbf{0.0635} \\
        \cmidrule{2-7} % 中间线
        & 7/32 & 18.87 & LLM-S[None]/ShortGPT & 103.59 & - & - \\
        & 7/32 & 18.65 & LINEARPATCH & \textbf{13.87} & \textbf{86.61} & 0.0211 \\
        & 7/32 & 18.87 & PCC & 21.21 & 79.53 & \textbf{4.9706} \\
        \bottomrule % 底部粗线
\end{tabular}
\caption{The performance of different pruning methods with or without compensation across different LLMs on the WikiText2 dataset of langauge modeling. $L_p$ and $L_t$ denotes the number of pruned layers and all layers in the model. Ratio denotes the pruning ratio (\%). Imp. represents their percentage improvement of PPL compared to pruning methods without compensation.}
\label{table4}
\end{table*}

\begin{table*}[h!]
    \centering
    \setlength{\tabcolsep}{1mm}
    \begin{tabular}{c c c c c c} % 去除竖线，改为四列左对齐
        \toprule % 顶部粗线
        Model & $L_p$/$L_t$ & Ratio & Method & PPL & Imp.(\%) \\
        \midrule % 中间线
        \multirow{9}{*}{\centering LLaMA3-8B} & 7/32 & 19.01 & ShortGPT & 57.77 & - \\
        & 7/32 & 18.80 & +LINEARPATCH & 25.35 & 56.12 \\
        & 7/32 & 18.80 & +LINEARPATCH+PCC & \textbf{24.48} & \textbf{57.52}  \\
        \cmidrule{2-6} % 中间线
        & 7/32 & 19.01 & LLM\_S[None] & 2305.17 & -  \\
        & 7/32 & 18.80 & +LINEARPATCH & 72.12 & 96.87 \\
        & 7/32 & 18.80 & +LINEARPATCH+PCC & \textbf{52.74} & \textbf{97.71}  \\
        \cmidrule{2-6} % 中间线
        & 7/32 & 19.01 & SLEB & 13.11 & - \\
        & 7/32 & 18.80 & +LINEARPATCH & 12.07 & 7.93 \\
        & 7/32 & 18.80 & +LINEARPATCH+PCC & \textbf{12.00} & \textbf{8.47}  \\
        \midrule % 中间线
        \multirow{6}{*}{\centering LLaMA2-7B} & 7/32 & 21.02 & ShortGPT/LLM\_S[None] & 18.45 & -  \\
        & 7/32 & 20.78 & +LINEARPATCH & 13.27 & 28.07  \\
        & 7/32 & 20.78 & +LINEARPATCH+PCC & \textbf{13.04} & \textbf{29.32}  \\
        \cmidrule{2-6} % 中间线
        & 7/32 & 21.02 & SLEB & 9.14 & - \\
        & 7/32 & 20.78 & +LINEARPATCH & 8.77 & 4.05 \\
        & 7/32 & 20.78 & +LINEARPATCH+PCC & \textbf{8.74} & \textbf{4.38}  \\
        \midrule % 中间线
        \multirow{6}{*}{\centering LLaMA2-13B} & 10/40 & 24.37 & ShortGPT/LLM\_S[None] & 9.77 & - \\
        & 10/40 & 24.17 & +LINEARPATCH & 8.74 & 10.54  \\
        & 10/40 & 24.17 & +LINEARPATCH+PCC & \textbf{8.67} & \textbf{11.26}  \\
        \cmidrule{2-6} % 中间线
        & 8/40 & 19.50 & SLEB & \textbf{8.17} & -  \\
        & 8/40 & 19.30 & +LINEARPATCH & 8.29 & -  \\
        & 8/40 & 19.30 & +LINEARPATCH+PCC & 8.27 & -  \\
        \bottomrule % 底部粗线
\end{tabular}
\caption{The performance of different pruning methods compensated by LINEARPATCH and PCC across different LLMs on the WikiText2 dataset of langauge modeling. $L_p$ and $L_t$ denotes the number of pruned layers and all layers in the model. Ratio denotes the pruning ratio (\%). Imp. represents their percentage improvement of PPL compared to pruning methods without compensation.}
\label{table10}
\end{table*}

\begin{table*}[t!]
    \centering
    \resizebox{\linewidth}{!}{
    \begin{tabular}{c c c c c c c c c c c c c} % 去除竖线，改为四列左对齐
        \toprule % 顶部粗线
        Model & Method & $\text{ARC}_c$ & $\text{ARC}_e$ & BoolQ & HeSw & PIQA & WG & WSC & MNLI & ANLI & Avg. & RP \\
        \midrule % 中间线
        \multirow{10}{*}{\centering LLaMA3-8B} & Dense & 53.49 & 77.73 & 81.28 & 79.15 & 80.95 & 73.16 & 69.23 & 47.96 & 35.40 & 66.48 & 100 \\
        & SliceGPT & 30.97 & 49.24 & 38.53 & 49.89 & 61.26 & 61.88 & 63.46 & 34.76 & 33.50 & 47.05 & 70.77 \\
        \cmidrule{2-13} % 中间线
        & SLEB & 34.47 & 60.01 & 44.49 & \textbf{62.04} & 73.94 & \textbf{55.25} & 59.61 & \textbf{34.78} & 32.71 & 50.81 & 76.43 \\
        & PCC & \textbf{34.90} & \textbf{60.35} & \textbf{44.74} & \textbf{62.04} & \textbf{74.10} & 54.85 & \textbf{61.54} & 34.25 & \textbf{33.41} & \textbf{51.13} & \textbf{76.91} \\
        \cmidrule{2-13} % 中间线
        & ShortGPT & 42.49 & 56.56 & 65.13 & 64.59 & 71.21 & \textbf{70.95} & 44.23 & 48.07 & 38.03 & 55.70 & 83.78 \\
        & LINEARPATCH & \textbf{43.34} & \textbf{60.61} & \textbf{76.70} & \textbf{67.01} & \textbf{73.28} & \textbf{70.95} & \textbf{66.35} & 50.34 & \textbf{39.16} & \textbf{60.86} & \textbf{91.55} \\
        & PCC & 42.41 & 59.42 & 75.32 & 64.52 & 71.15 & 70.32 & 64.42 & \textbf{51.42} & 36.78 & 59.53 & 89.55 \\
        \cmidrule{2-13} % 中间线
        & LLM\_S[None] & 29.10 & 39.77 & 38.04 & 33.20 & 58.92 & 55.95 & \textbf{63.46} & 33.28 & 33.84 & 42.84 & 64.44 \\
        & LINEARPATCH & 34.47 & 49.62 & \textbf{55.44} & 48.95 & 63.38 & 63.53 & \textbf{63.46} & \textbf{46.20} & 37.21 & \textbf{51.36} & \textbf{77.26} \\
        & PCC & \textbf{36.52} & \textbf{50.55} & 38.44 & \textbf{51.68} & \textbf{64.14} & \textbf{68.27} & \textbf{63.46} & 45.04 & \textbf{37.47} & 50.62 & 76.14 \\
        \midrule % 中间线
        \multirow{8}{*}{\centering LLaMA2-7B} & Dense & 46.41 & 74.62 & 77.67 & 75.99 & 79.10 & 69.29 & 36.53 & 42.23 & 37.15 & 59.89 & 100 \\
        & SliceGPT & 35.24 & 58.00 & 48.44 & 56.93 & 69.15 & 63.69 & 63.46 & 35.09 & 35.33 & 51.70 & 86.32 \\
        \cmidrule{2-13} % 中间线
        & SLEB & \textbf{33.10} & 56.40 & 64.04 & 62.48 & 73.12 & \textbf{58.96} & \textbf{36.53} & 40.88 & 33.81 & 51.04 & 85.22 \\
        & PCC & 33.02 & \textbf{56.82} & \textbf{65.23} & \textbf{62.50} & \textbf{73.29} & \textbf{58.96} & \textbf{36.53} & \textbf{41.11} & \textbf{33.95} & \textbf{51.27} & \textbf{85.61} \\
        \cmidrule{2-13} % 中间线
        & ShortGPT & 36.26 & 55.89 & \textbf{62.17} & 62.63 & 70.40 & 66.21 & \textbf{36.53} & 35.64 & \textbf{34.31} & 51.12 & 85.36 \\
        & LLM\_S[None] & 36.26 & 55.89 & \textbf{62.17} & 62.63 & 70.40 & 66.21 & \textbf{36.53} & 35.64 & \textbf{34.31} & 51.12 & 85.36 \\
        & LINEARPATCH & \textbf{37.71} & \textbf{61.23} & 62.14 & \textbf{63.64} & \textbf{71.38} & 65.66 & \textbf{36.53} & \textbf{37.04} & 32.34 & \textbf{51.96} & \textbf{86.76} \\
        & PCC & 36.77 & 58.12 & \textbf{62.17} & 63.13 & 69.91 & \textbf{66.54} & \textbf{36.53} & 35.70 & 32.90 & 51.31 & 85.67 \\
        \midrule % 中间线
        \multirow{8}{*}{\centering LLaMA2-13B} & Dense & 49.06 & 77.52 & 80.58 & 79.35 & 80.52 & 72.13 & 44.23 & 43.11 & 38.25 & 62.75 & 100 \\
        & SliceGPT & 40.87 & 61.45 & 40.91 & 57.46 & 68.49 & 68.03 & 63.46 & 35.05 & 32.30 & 52.00 & 82.87 \\
        \cmidrule{2-13} % 中间线
        & SLEB & 36.43 & \textbf{61.86} & \textbf{62.32} & \textbf{67.02} & 75.02 & 62.35 & \textbf{36.53} & 34.92 & \textbf{33.95} & 52.27 & 83.30 \\
        & PCC & \textbf{36.60} & \textbf{61.86} & \textbf{62.32} & 66.97 & \textbf{75.03} & \textbf{62.67} & \textbf{36.53} & \textbf{35.04} & 33.93 & \textbf{52.33} & \textbf{83.40} \\
        \cmidrule{2-13} % 中间线
        & ShortGPT & 42.92 & 63.55 & 58.22 & 69.26 & 72.57 & 69.69 & \textbf{39.42} & 42.87 & 37.40 & 55.10 & 87.81 \\
        & LLM\_S[None] & 42.92 & 63.55 & 58.22 & 69.26 & 72.57 & 69.69 & \textbf{39.42} & 42.87 & 37.40 & 55.10 & 87.81 \\
        & LINEARPATCH & \textbf{43.68} & \textbf{65.44} & 62.08 & \textbf{70.43} & \textbf{74.15} & 70.32 & 36.53 & \textbf{46.93} & 37.18 & \textbf{56.30} & \textbf{89.72} \\
        & PCC & 43.09 & 64.81 & \textbf{62.48} & 69.23 & 72.85 & \textbf{70.80} & 36.53 & 46.02 & \textbf{39.69} & 56.17 & 89.51 \\
        \midrule % 中间线
        \multirow{8}{*}{\centering Baichuan2-7B} & Dense & 42.74 & 72.93 & 73.12 & 72.28 & 77.42 & 67.56 & 57.69 & 42.61 & 38.25 & 62.75 & 100 \\
        \cmidrule{2-13} % 中间线
        & SLEB & 29.86 & 55.68 & 61.07 & \textbf{55.80} & \textbf{71.93} & 55.48 & \textbf{51.92} & 35.74 & 34.44 & \textbf{50.21} & \textbf{80.02} \\
        & PCC & \textbf{29.95} & \textbf{55.98} & \textbf{61.56} & 55.22 & 71.65 & \textbf{55.96} & 50.00 & \textbf{35.99} & \textbf{34.69} & 50.11 & 79.86 \\
        \cmidrule{2-13} % 中间线
        & ShortGPT & \textbf{35.06} & 51.72 & 62.38 & 55.32 & 65.56 & 64.48 & \textbf{36.53} & 41.80 & 38.78 & 50.18 & 79.97 \\
        & LLM\_S[None] & \textbf{35.06} & 51.72 & 62.38 & 55.32 & 65.56 & 64.48 & \textbf{36.53} & 41.80 & 38.78 & 50.18 & 79.97 \\
        & LINEARPATCH & 34.72 & 56.86 & 62.75 & \textbf{58.67} & \textbf{68.66} & \textbf{66.38} & \textbf{36.53} & 38.82 & 36.72 & 51.12 & 81.47 \\
        & PCC & 33.96 & \textbf{56.95} & \textbf{65.32} & 56.70 & 67.57 & 65.59 & \textbf{36.53} & \textbf{42.79} & \textbf{39.03} & \textbf{51.61} & \textbf{82.25} \\
        \bottomrule % 底部粗线
\end{tabular}
}
\caption{The performance of different pruning methods with or without compensation across different LLMs on the 9 mainstream datasets of zero-shot QA.
Avg. denotes the average accuracy (\%). RP represents the retained performance relative to the original model.}
\label{table5}
\end{table*}

\begin{table*}[t!]
    \centering
    \resizebox{\linewidth}{!}{
    \begin{tabular}{c c c c c c c c c c c c} % 去除竖线，改为四列左对齐
        \toprule % 顶部粗线
        Model & Method & $\text{ARC}_c$ & $\text{ARC}_e$ & BoolQ & HeSw & PIQA & WG & WSC & MNLI & ANLI & Avg.  \\
        \midrule % 中间线
        \multirow{6}{*}{\centering LLaMA3-8B} & ShortGPT & 42.49 & 56.56 & 65.13 & 64.59 & 71.21 & \textbf{70.95} & 44.23 & 48.07 & 38.03 & 55.70  \\
        & +LINEARPATCH & \textbf{43.34} & 60.61 & \textbf{76.70} & 67.01 & \textbf{73.28} & \textbf{70.95} & \textbf{66.35} & 50.34 & \textbf{39.16} & 60.86  \\
        & +LINEARPATCH+PCC & 43.17 & \textbf{61.57} & 76.27 & \textbf{67.25} & 73.01 & 70.80 & \textbf{66.35} & \textbf{50.42} & 39.00 & \textbf{60.87}  \\
        \cmidrule{2-12}
        & LLM\_S[None] & 29.10 & 39.77 & 38.04 & 33.20 & 58.92 & 55.95 & \textbf{63.46} & 33.28 & 33.84 & 42.84  \\
        & +LINEARPATCH & 34.47 & 49.62 & \textbf{55.44} & 48.95 & 63.38 & 63.53 & \textbf{63.46} & \textbf{46.20} & 37.21 & 51.36  \\
        & +LINEARPATCH+PCC & \textbf{35.75} & \textbf{53.87} & 55.17 & \textbf{52.69} & \textbf{65.07} & \textbf{65.04} & \textbf{63.46} & 45.63 & \textbf{37.25} & \textbf{52.66}  \\
        \midrule
        \multirow{3}{*}{\centering LLaMA2-7B} & ShortGPT/LLM\_S[None] & 36.26 & 55.89 & \textbf{62.17} & 62.63 & 70.40 & 66.21 & \textbf{36.53} & 35.64 & 34.31 & 51.12  \\
        & +LINEARPATCH & 37.71 & \textbf{61.23} & 62.14 & \textbf{63.64} & \textbf{71.38} & 65.66 & \textbf{36.53} & 37.04 & 32.34 & 51.96 \\
        & +LINEARPATCH+PCC & \textbf{38.31} & 60.90 & 62.14 & 63.54 & 71.11 & \textbf{66.07} & \textbf{36.53} & \textbf{37.40} & \textbf{32.41} & \textbf{52.05} \\
        \midrule
        \multirow{3}{*}{\centering LLaMA2-13B} & ShortGPT/LLM\_S[None] & 42.92 & 63.55 & 58.22 & 69.26 & 72.57 & 69.69 & \textbf{39.42} & 42.87 & 37.40 & 55.10  \\
        & +LINEARPATCH & \textbf{43.68} & 65.44 & 62.08 & 70.43 & 74.15 & \textbf{70.32} & 36.53 & 46.93 & \textbf{37.18} & \textbf{56.30} \\
        & +LINEARPATCH+PCC & 43.34 & \textbf{65.57} & \textbf{62.29} & \textbf{70.44} & \textbf{74.37} & 70.17 & 36.53 & \textbf{47.07} & 36.95 & \textbf{56.30} \\
        \bottomrule % 底部粗线
\end{tabular}
}
\caption{The performance of different pruning methods compensated by LINEARPATCH and PCC across different LLMs on the 9 mainstream datasets of zero-shot QA. Avg. denotes the average accuracy (\%).}
\label{table21}
\end{table*}

Our experimental results in Table~\ref{table4} show that, for the language modeling task, our PCC method achieves an average PPL improvement of 3.58\% and 52.23\% for the cross-layer and consecutive-layer pruning methods. PCC achieves 94\% of the average performance of LINEARPATCH, which is the SOTA method for full-channel compensation, by merely compensating 0.5\% of the channels. Furthermore, PCC outperforms LINEARPATCH on both the LLaMA3-8B and LLaMA2-13B models. For the CAPI metric, our PCC method shows a substantial performance advantage over LINEARPATCH.

Compared to LINEARPATCH, our PCC method introduces much fewer additional parameters. LINEARPATCH applies Hadamard transforms to each pruning interface, thus necessitating the additional storage of two Hadamard matrices per interface.

Experimental results in Table~\ref{table10} show that as a plug-and-play method, PCC can be seamlessly integrated into LINEARPATCH. PCC further improves performance by 0.44\% and 1.08\% for cross-layer pruning and consecutive-layer pruning, respectively, achieving SOTA results on language modeling tasks. This strongly validates that pivotal channels are critical to preserving and enhancing model performance.

Notably, applying LINEARPATCH compensation to LLaMA2-13B after cross-layer pruning degrades the model’s performance. We attribute this to the excessive number of pruning interfaces requiring compensation. However, the LINEARPATCH method fails to distinguish pivotal channels from other channels, as compensation is performed via a uniform scaling operation—leading to error accumulation across the 6 pruning interfaces. In contrast, our PCC method focuses on precise compensation of pivotal channels. As demonstrated in Table~\ref{table4}, the PCC method does not introduce adverse effects on the compensated model.

\subsubsection{Results on Zeor-Shot Question Answering}

Table~\ref{table5} show that in zero-shot tasks, the performance improvement following consecutive-layer pruning is significantly superior to that after cross-layer pruning, and this discrepancy may be attributed to inter-layer differences in channel importance and semantic representation, which pose challenges for cross-layer compensation. Compared with the full-channel compensation method LINEARPATCH, PCC only compensates 0.5\% of all channels while achieving highly competitive compensation performance, and it even surpasses LINEARPATCH on the Baichuan2-7B model. Although LINEARPATCH compensates all channels and theoretically enjoys a higher performance ceiling, PCC still matches or exceeds LINEARPATCH on various tasks for other models. We observe that the performance improvements of the PCC compensated model on the MNLI,ANLI, and BoolQ tasks are generally more significant and broader than those on the ARC tasks. This can be attributed to the fact that MNLI, ANLI, and BoolQ, which emphasize textual logical relationships, impose demands on the semantic comprehension and fine-grained semantic association modeling capabilities of the model. Our PCC method effectively alleviates misalignments in the original non-consecutive layer vector space by selectively compensating for channels with significant changes at the pruning interface, thereby reducing the skip propagation of semantic dependencies, which aligns with the demands of the aforementioned high-semantic-sensitivity tasks. ARC, by contrast, focuses on scientific knowledge reasoning tasks, where compensating for such long-tail information is more complex, typically requiring retraining or compensation for more general channels. Our experimental results support this conclusion. After excluding the scientific reasoning task ARC\_Easy, ARC\_Challenge, and the physical commonsense reasoning task PIQA, the performance of PCC and LINEARPATCH are comparable on LLaMA2-7B. In contrast, PCC outperforms LINEARPATCH on LLaMA2-13B.

To further explore the potential of PCC, in Table~\ref{table21} we integrate PCC into LINEARPATCH on LLaMA2-7B/13B and LLaMA3-8B. Since PCC already outperforms LINEARPATCH on the Baichuan2-7B model, we did not conduct further extended experiments on this model. Experimental results verify that targeted compensation for distinct pivotal channels yields steady performance gains across diverse models, demonstrating the strong generalization ability of PCC. We would like to clarify that this combined experiment serves only as an initial attempt rather than the optimal integration of the two methods. LINEARPATCH leverages Hadamard transformation to smooth activations for subsequent scaling and alignment, while PCC centers on accurately locating and rectifying pivotal channels. Their underlying mechanisms are partially incompatible: the Hadamard transformation in LINEARPATCH distorts the inherent distribution features of pivotal channels and accordingly impairs PCC’s performance improvement to some extent. Even under such unfavorable constraints on PCC’s performance, the combined experiment still highlights PCC’s outstanding expandability and practical potential.

\paragraph{Discussion 3: Advantages, Limitations, and Internal Connections of two methods} As a lightweight plug-and-play module, PCC only compensates 0.5\% of the total channels to achieve performance comparable to or even better than full-channel compensation, with low training cost and few extra parameters, and it can be readily integrated with other compensation methods. Built on the observation that pivotal channels have higher Pearson correlation coefficients, the proposed PCC adopts learnable linear transformations for compensation. PCC theoretically solidifies the compensation strategy and first reveals that more compensated channels do not always yield better results, as compensating certain channels may instead impair model performance. LINEARPATCH, by contrast, mitigates the adverse interference caused by outliers during full-channel compensation and further improves the efficacy of full-channel compensation. A shared limitation of the two approaches is that neither takes the interaction effects among channels on compensation outcomes into consideration. Both methods fall into the category of post-pruning channel compensation and share several intrinsic connections. (1) Both methods share the primary purpose of repairing semantic discontinuities caused by layer pruning via linear transformations on output hidden states at pruning interface. (2) PCC constitutes an evolutionary improvement over LINEARPATCH, the coarse-grained full-channel compensation method. Building upon such coarse-grained compensation, PCC proposes a fine-grained targeted compensation method that only optimizes pivotal channels. (3) PCC and LINEARPATCH are not mutually exclusive and can be integrated. By adjusting only pivotal channels and keeping the inherent features of ordinary channels intact, PCC works as a favorable plug-and-play module that can be combined with alternative channel compensation algorithms, indicating promising research prospects. Experimental results demonstrate that the hybrid of PCC and LINEARPATCH achieves prominent performance improvements even under non-optimal matching configurations. In future work, we will focus on exploring efficient integration paradigms for PCC and full-channel compensation techniques represented by LINEARPATCH. We intend to redistribute and align output activation magnitudes while preserving the intrinsic feature information of pivotal channels as much as possible. This design enables these channels to be seamlessly incorporated into the PCC compensation framework for further prominent performance improvements.

\subsection{Ablation Studies}
\subsubsection{Rationale for Adopting Variance as the Sole Metric for Pivotal Channel Selection}

Existing studies on channel/layer pruning and post-pruning compensation typically rely on single-layer activation magnitude to evaluate channel importance; for instance, RECAP \cite{lee2025recap} prunes channels according to average activation values. For layer pruning compensation, it is widely recognized that performance degradation mainly stems from semantic gaps caused by concatenating original unconnected layers. Aligning activation magnitudes serves as a straightforward remedy. Prior studies still rely heavily on activation values: Sun et al. \cite{DBLP:journals/corr/abs-2402-17762} observe 3-4 channels with remarkably high activations in intermediate layers of LLMs, while LINEARPATCH \cite{DBLP:conf/nips/ChenBYLZYHGHY25} smooths such high-magnitude channels via Hadamard transform, as these channels hinder post-pruning feature alignment.

\begin{figure*}[t]
\centering
\includegraphics[width=0.9\textwidth]{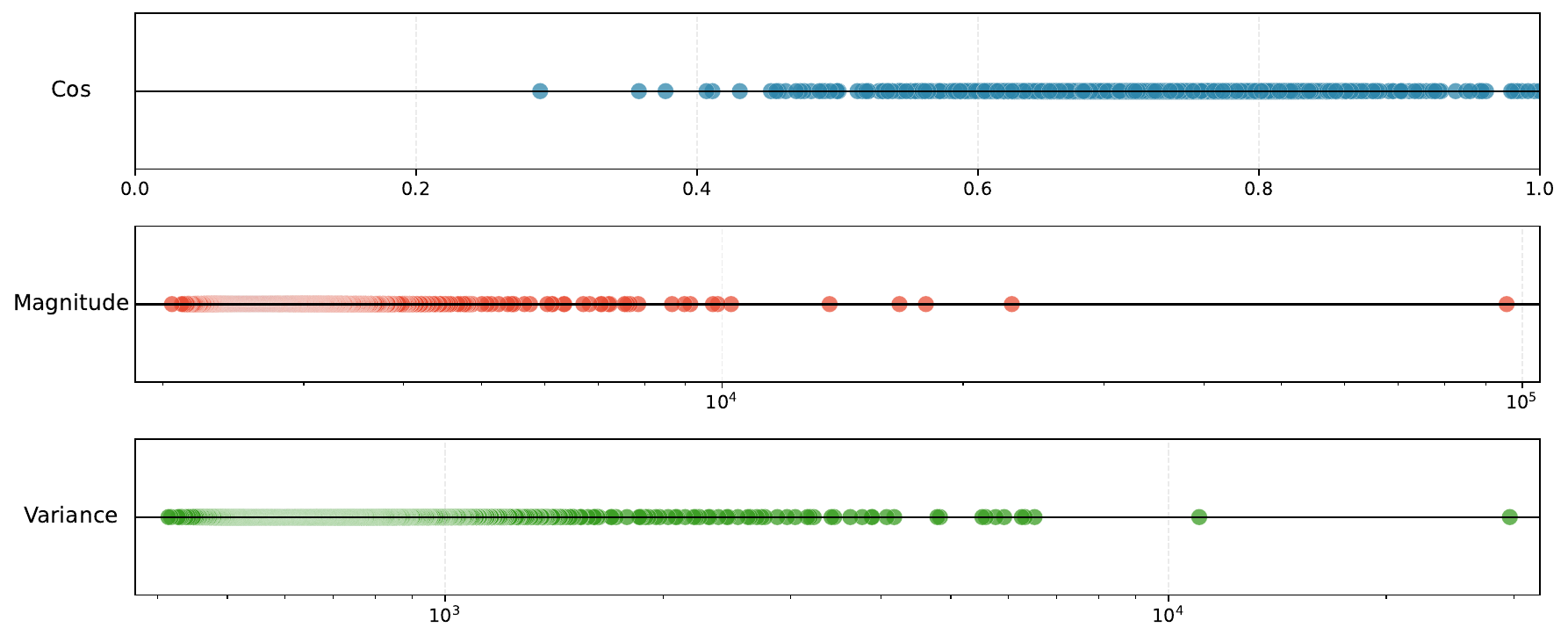} % Reduce the figure size so that it is slightly narrower than the column.
\caption{The score distribution of various evaluation metrics for pivotal channels.}
\label{fig7}
\end{figure*}

Nevertheless, we argue that the semantic gap originates from shifted feature distributions and broken semantic dependencies between unconnected layers at pruning interface rather than abnormal single-layer activations. Unlike activation magnitude, which only reflects the single-layer response intensity and fails to describe cross-layer feature variation, variance and cosine similarity can precisely quantify the cross-layer fluctuation of channel features around pruning positions, making it more appropriate for offsetting pruning-caused semantic discrepancy. However, our experiments reveal that using cosine similarity to select pivotal channels has notable limitations. As illustrated in Figure~\ref{fig7}, activation magnitude and variance range from 0 to their respective maximum values. For models such as LLaMA2-7B with 4096 channels, these two metrics yield widely distributed results with strong discriminative power. In contrast, cosine similarity is restricted to the interval of 0-1, making it impractical to select pivotal channels solely based on this metric.

Accordingly, instead of following the prevalent paradigm based on single-layer activation magnitude or cosine similarity, we adopt inter-layer variance to screen pivotal channels with the strongest information representation capability. Variance directly reflects the cross-layer variation of channel features; channels with more fluctuations carry more critical semantic information, so we select variance as our criterion for determining the pivotal channels. To verify the rationality, we supplement extra ablation experiments. Taking LLaMA2-7B as an example and fixing the total number of compensated channels, we compare the compensation performance between high-activation channels and high-variance channels. As shown in Table~\ref{table20}, compensating high-variance channels achieves better performance restoration than compensating channels with high-activation values.

\begin{table*}[t]
    \centering
    \begin{tabular}{c c c c} % 去除竖线，改为四列左对齐
        \toprule % 顶部粗线
        Method & Origin & After Compensation & Imp.(\%)  \\
        \midrule % 中间线
        Magnitude & 18.45 & 15.23 & 17.45 \\
        Variance & 18.45 & \textbf{14.50} & \textbf{21.41} \\
        \bottomrule % 底部粗线
\end{tabular}
\caption{The ablation study on various evaluation metrics for pivotal channels.}
\label{table20}
\end{table*}

\subsubsection{The Importance of Pivotal Channels}
To investigate the impact of two types of channels, i.e., pivotal channels and other channels, on model performance, we disrupt these two types of channels at the pruning interface in two ways, including setting the channel activation values to 0 ($C_0$) and amplifying them by 10 times ($C_{\times10}$). Table~\ref{table6} presents the PPL of the perturbed LLaMA2-7B model on WikiText2 dataset for language modeling.

\begin{table*}[t]
    \centering
    \begin{tabular}{c c c c} % 去除竖线，改为四列左对齐
        \toprule % 顶部粗线
        State & Channel & $C_0$ & $C_{\times10}$  \\
        \midrule % 中间线
        Dense & - & 9.14 & 9.14 \\
        \midrule % 中间线
        \multirow{5}{*}{\centering Pivotal} & 1512 & 9.19 & 18.17 \\
        & 2158 & 9.38 & 31.93 \\
        & 2350 & 9.33 & 17.05 \\
        & 2298 & 9.19 & 25.92 \\
        & 2235 & 9.27 & 93.55 \\
        \midrule % 中间线
        \multirow{5}{*}{\centering Other} & 151 & 9.14 & 9.18 \\
        & 215 & 9.14 & 9.18 \\
        & 235 & 9.14 & 9.16 \\
        & 229 & 9.14 & 9.19 \\
        & 223 & 9.14 & 9.18 \\
        \bottomrule % 底部粗线
\end{tabular}
\caption{The PPL of the perturbed LLaMA2-7B model on WikiText2 dataset for language modeling.}
\label{table6}
\end{table*}

Experimental results show that pivotal channels exert a critical influence on model performance. Specifically, human intervention that disrupts the normal expression of pivotal channels can cause significant decrease in model performance. In contrast, the impact on other channels is negligible. This discovery highlights the importance of pivotal channels.

\begin{table*}[t!]
    \centering
    \begin{tabular}{c c c c} % 去除竖线，改为四列左对齐
        \toprule % 顶部粗线
        Model & Layer & Type & PPL After Compensation  \\
        \midrule % 中间线
        \multirow{9}{*}{\centering LLaMA2-7B} &   - & SLEB Pruning & 9.14 \\
        \cmidrule{2-4}
        & \multirow{2}{*}{\centering 9} & Pivotal  & 9.0743 \\
        & & Other  & 9.1401 \\
        \cmidrule{2-4}
        & \multirow{2}{*}{\centering 13} & Pivotal  & 9.0332  \\
        & & Other  & 9.1398 \\
        \cmidrule{2-4}
        & \multirow{2}{*}{\centering 22} & Pivotal  & 8.9315  \\
        & & Other  & 9.1394 \\
        \cmidrule{2-4}
        & \multirow{2}{*}{\centering 26} & Pivotal  & 8.9125  \\
        & & Other  & 9.1393 \\
        \midrule % 中间线
        \multirow{14}{*}{\centering LLaMA2-13B} & - & SLEB Pruning & 8.17 \\
        \cmidrule{2-4}
        & \multirow{2}{*}{\centering 2} & Pivotal  & 8.0085  \\
        & & Other  & 8.1706 \\
        \cmidrule{2-4}
        & \multirow{2}{*}{\centering 12,13,14} & Pivotal  & 7.9192  \\
        & & Other  & 8.1697 \\
        \cmidrule{2-4}
        & \multirow{2}{*}{\centering 27} & Pivotal  & 7.9168  \\
        & & Other  & 8.1698 \\
        \cmidrule{2-4}
        & \multirow{2}{*}{\centering 29} & Pivotal  & 7.9159  \\
        & & Other  & 8.1701 \\
        \cmidrule{2-4}
        & \multirow{2}{*}{\centering 31} & Pivotal  & 7.9130  \\
        & & Other  & 8.1702 \\
        \cmidrule{2-4}
        & \multirow{2}{*}{\centering 33} & Pivotal  & 7.9079  \\
        & & Other  & 8.1702 \\
        \midrule % 中间线
        \multirow{7}{*}{\centering LLaMA3-8B} & - & SLEB Pruning & 13.11 \\
        \cmidrule{2-4}
        & \multirow{2}{*}{\centering 9,10,11,12} & Pivotal  & 12.6713  \\
        & & Other  & 13.1182 \\
        \cmidrule{2-4}
        & \multirow{2}{*}{\centering 23} & Pivotal  & 12.5571  \\
        & & Other  & 13.1173 \\
        \cmidrule{2-4}
        & \multirow{2}{*}{\centering 25,26} & Pivotal  & 12.4280  \\
        & & Other  & 13.1123 \\
        \bottomrule % 底部粗线
\end{tabular}
\caption{The discrepancy of compensation for different channels.}
\label{table7}
\end{table*}

\begin{figure*}[h!]
    \centering
    % 第一行子图
    \begin{subfigure}{0.45\linewidth}
        \centering
        \includegraphics[width=\linewidth]{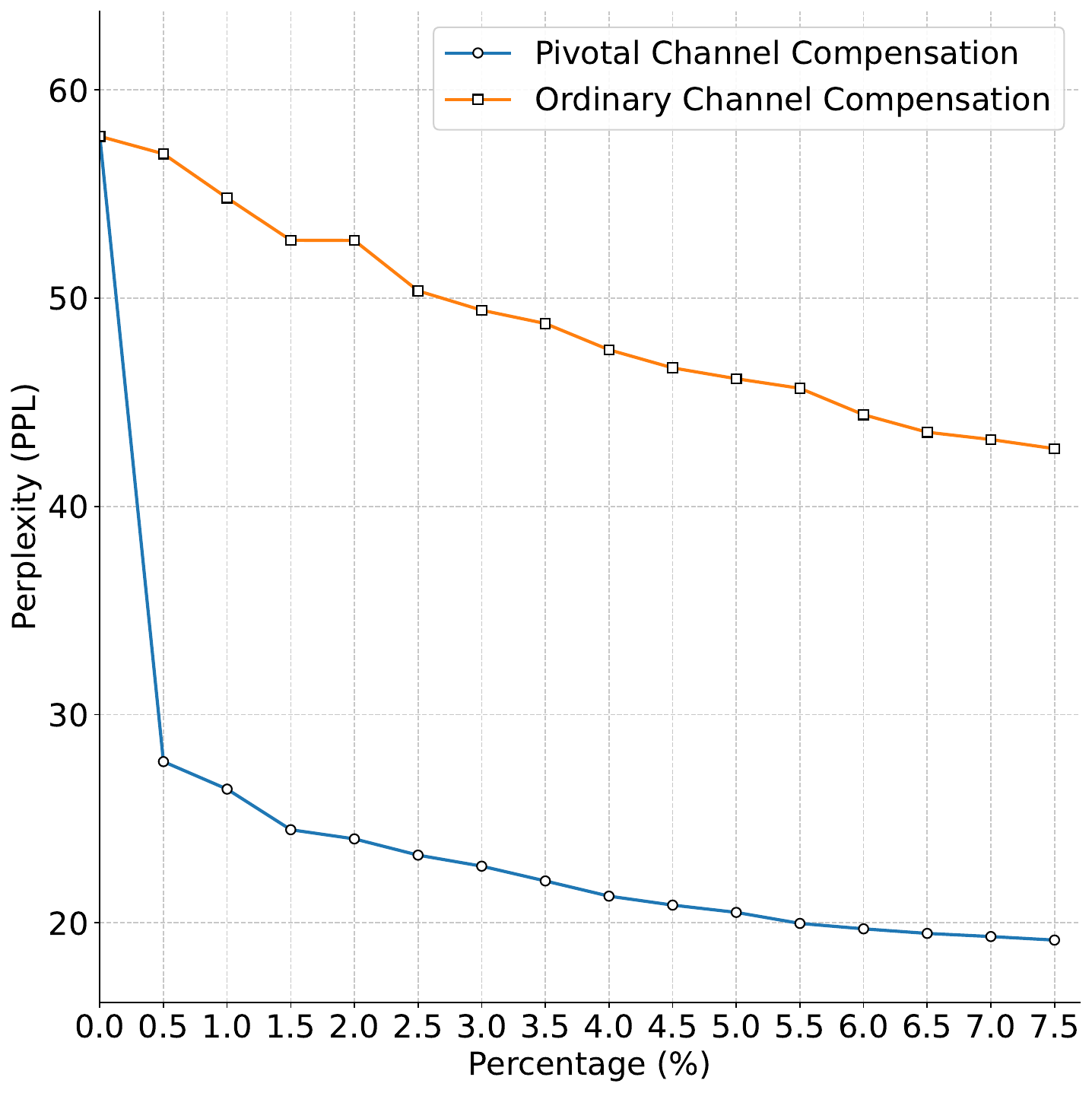}
        \caption{The PPL across different compensation rates (the proportions of pivotal channels) for LLaMA3-8B.}
        \label{fig:sub17}
    \end{subfigure}
    \hfill
    \begin{subfigure}{0.45\linewidth}
        \centering
        \includegraphics[width=\linewidth]{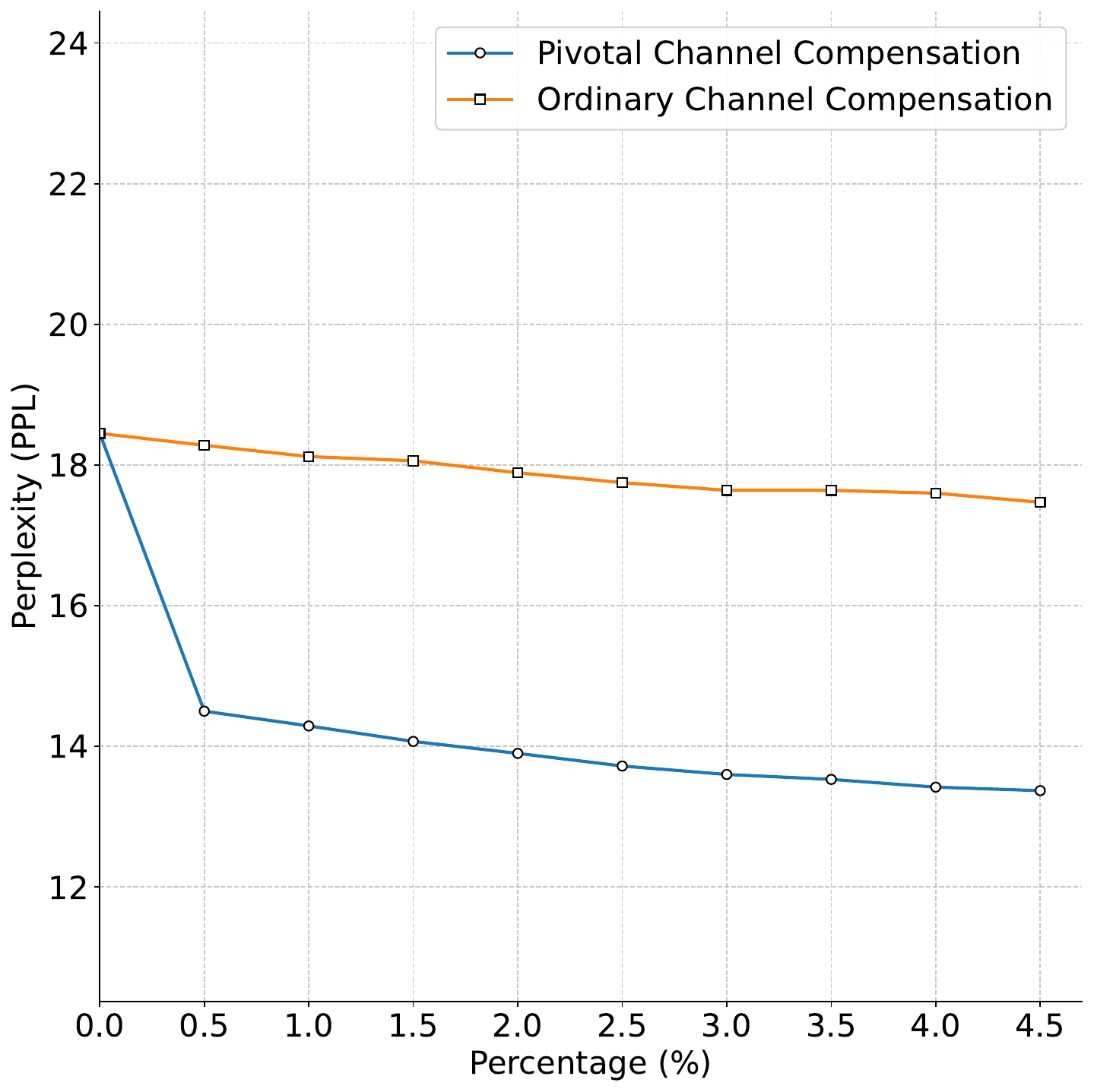}
        \caption{The PPL across different compensation rates (the proportions of pivotal channels) for LLaMA2-7B.}
        \label{fig:sub16}
    \end{subfigure}
    \smallskip % 中等大小的垂直间距（比\vspace更符合LaTeX排版习惯）
    % 第二行子图
    \begin{subfigure}{0.45\linewidth}
        \centering
        \includegraphics[width=\linewidth]{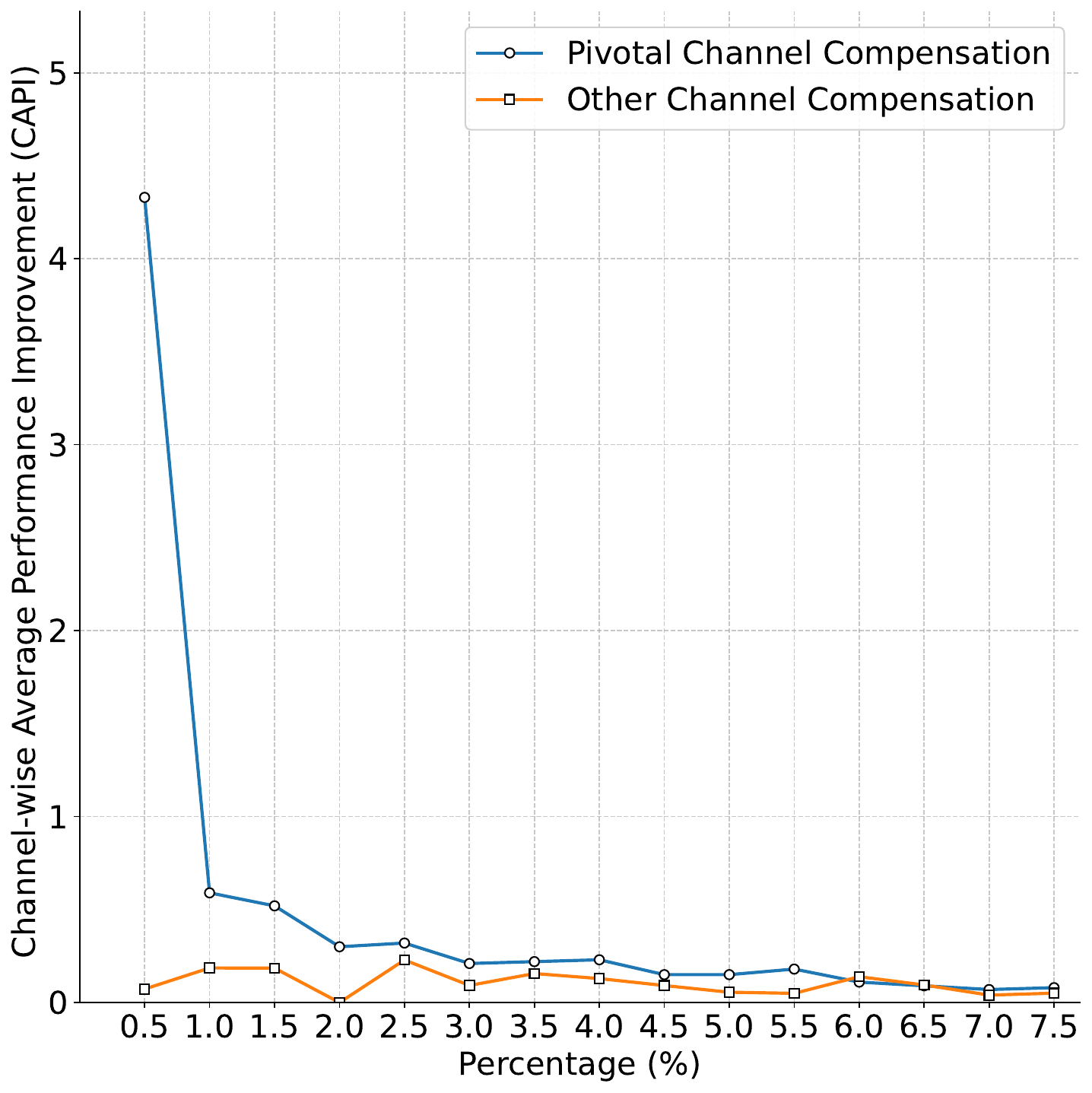}
        \caption{The CAPI across different compensation rates (the proportions of pivotal channels) for LLaMA3-8B.}
        \label{fig:sub18}
    \end{subfigure}
    \hfill
    \begin{subfigure}{0.45\linewidth}
        \centering
        \includegraphics[width=\linewidth]{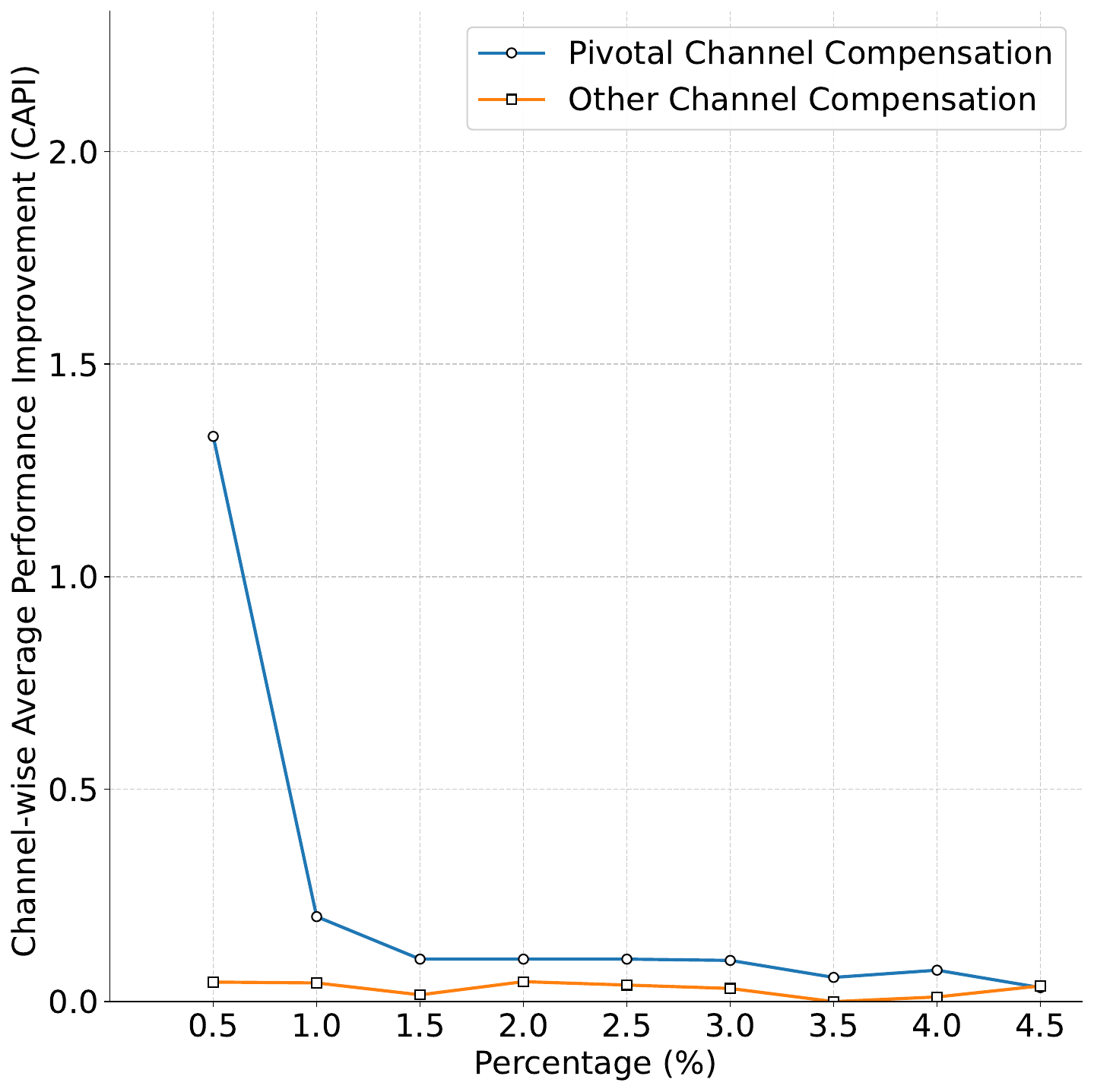}
        \caption{The CAPI across different compensation rates (the proportions of pivotal channels) for LLaMA2-7B.}
        \label{fig:sub19}
    \end{subfigure}
    \caption{The performance across different compensation rates (the proportions of pivotal channels) for different LLMs.}
    \label{fig6}
  \end{figure*}

\subsubsection{Ablation Study of PCC}
Taking the cross-layer pruning as an example, we compare the performance improvements by compensating pivotal channels and other channels respectively. We use the same compensation method as PCC for other channels that are equal in number to the pivotal channels.

As shown in Table~\ref{table7}, experimental results show that compensating for pivotal channels can bring greater performance improvements compared to compensating for other channels after pruning, which demonstrates the validity and efficiency of PCC.

\subsubsection{Ablation Study on Proportion of Pivotal Channels}
Previous works that compensate for all channels at the pruning interface is based on an intuitive assumption that the number of compensation channels is positively correlated with performance improvement. However, there are still two challenges: 1) the cost for compensating for thousands of channels is excessively high; 2) Uncertainty remains regarding whether the aforementioned assumption holds, and whether there exists a proportional relationship between channel compensation and performance improvement. Neither of these two issues has been addressed in previous studies. 

As shown in Figure~\ref{fig6}, across different models, compensating for 0.5\% of pivotal channels achieves significantly higher performance improvement than compensating other channels. Furthermore, compensating approximately 2.0\% of pivotal channels still results in a noticeable performance gain compared to other channel compensation. When the compensation ratio increases to 5.0\%, the performance improvement between the two channel types becomes nearly identical. To highlight the significant performance gain achieved by compensating a minimal set of pivotal channels, we set the compensation ratio to 0.5\%.

Regarding the second issue, we have found that not all channels can improve model performance when compensated. Taking the LLaMA2-7B model as an example: experiments on the calibration set show that linear compensation for two of the pivotal channels actually decreases the model’s language modeling performance, and this number proves to be even higher in other models. This is related to the selection of the calibration set and the compensation method.

\subsubsection{Ablation Study of Different Compensation Methods}
Previous full-channel compensation works all rely on scaling operations to compensate for semantic dependencies. However, channel compensation demands a high degree of refinement; otherwise, it may lead to a degradation in model performance. In contrast, our proposed PCC method focuses on pivotal channels, a tiny subset of all channels, which enables highly refined compensation for pivotal channels. Replacing traditional scaling operations with trainable linear compensation, the PCC method can more accurately restore the variations in channels. Taking the LLaMA3-8B model as an example, our experimental results demonstrate that the PPL of the model increases from 6.13 to 57.77 after ShortGPT layer pruning. When applying trainable linear compensation to 0.5\% of all channels, the PPL is restored to 27.74, achieving a performance improvement of 51.98\%. In contrast, using scaling-based compensation for the same proportion of channels only restores the PPL to 33.53, resulting in a performance improvement of 41.95\%.

\section{Discussion}
 To further explore the theoretical performance limit of the PCC method in the language modeling task for cross-layer pruning, we replace the activation values of pivotal channels at the pruning interface with those of the dense model during evaluation, thereby simulating a scenario of perfect compensation. 
Table~\ref{table8} shows that the compensation effects of our PCC method are already close to the theoretical optimal performance. Notably, this theoretical performance bound is still likely to be broken, as it only considers the impact of individual channels on model performance. We argue that the weaker compensation effects on cross-layer pruning are because we merely consider the impact of individual channels, without considering the influence of the interactions between pivotal channels on model performance. Therefore, in future, we will focus on the influence of the interactions between pivotal channels. Furthermore, we will also explore schemes for calibration-set-free channel output alignment to enhance the generalization capability of PCC across diverse datasets, given its limited performance improvement on long-tailed knowledge datasets.

While our proposed PCC method leverages variance as the sole indicator for pivotal channel screening, we have verified that variance serves as a highly effective standalone metric for this task. Even so, we firmly believe that multi-metric selection schemes represent an extremely promising research direction. The reasons why we did not adopt multi-metric fusion in the current work are elaborated as follows: (1) Multi-metric fusion involves weight allocation, introducing additional hyperparameters and reducing interpretability. (2) We argue that combining multiple metrics may achieve better performance. However, constrained by limited research efforts, we have not established a mature multi-metric evaluation framework. Accordingly, we will thoroughly investigate multi-metric fusion strategies for pivotal channel selection as a key direction in our future research.

\begin{table*}[t]
    \centering
    \begin{tabular}{c c c} % 去除竖线，改为四列左对齐
        \toprule % 顶部粗线
        Model &  Theoretical Optimum & PCC \\
        \midrule % 中间线
        LLaMA2-7B & 8.86 & 8.91 \\
        LLaMA2-13B & 7.71 & 7.9 \\
        LLaMA3-8B & 12.1 & 12.42 \\
        \bottomrule % 底部粗线
\end{tabular}
\caption{The theoretical performance bound of linear transformation of PCC.}
\label{table8}
\end{table*}

\section{Conclusion}
In this paper, we propose PCC, an efficient compensation method for layer pruning, after observing that the expression of a large proportion of information is concentrated on a small number of pivotal channels. This method addresses the key limitations of current mainstream compensation methods, including high training cost, excessive additional parameters, lack of interpretability, and difficulties in cross-layer compensation. We first identify pivotal channels with significant hidden state variations in inter-layer outputs based on variance, and selectively compensate for these pivotal channels via learnable linear transformations at the pruning interface. Extensive experimental results demonstrate that the PCC method achieves performance improvements that are nearly equivalent to or exceed those from compensating all channels, by compensating about 0.5\% of all channels, while introducing only a minimal number of additional parameters. We anticipate that our proposed PCC method will offer novel insights for efficient compensation of models after layer pruning.

\section*{Acknowledgments}
This work was supported by the National Natural Science Foundation of China (Grant No. 62406003) and the Anhui Provincial Scientific Research Plan Preparation Project (Grant No. 2024AH052212).

\bibliographystyle{unsrtnat}
\bibliography{references}  %%% Uncomment this line and comment out the ``thebibliography'' section below to use the external .bib file (using bibtex) .
\newpage

\appendix
\section{Supplementary Analysis on the Occurrence Frequency of Pivotal Channels}
\label{app1}
\begin{figure}[ht]
\centering
% 第一行子图
\begin{subfigure}{0.48\linewidth}
    \centering
    \includegraphics[width=\linewidth]{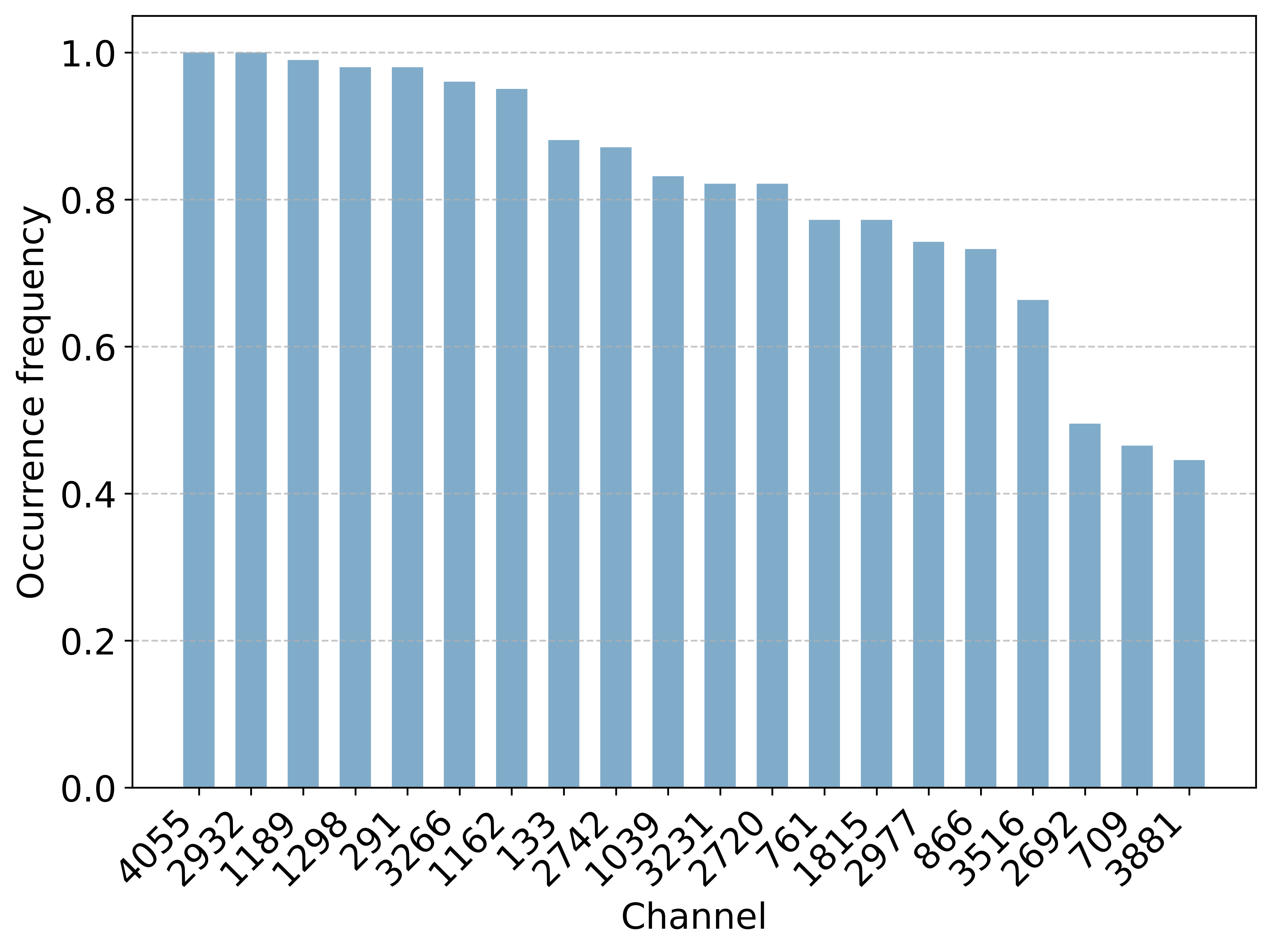}
    \caption{The occurrence frequency of pivotal channels between layer8 and layer12.}
    \label{fig:sub13}
\end{subfigure}
\hfill
\begin{subfigure}{0.48\linewidth}
    \centering
    \includegraphics[width=\linewidth]{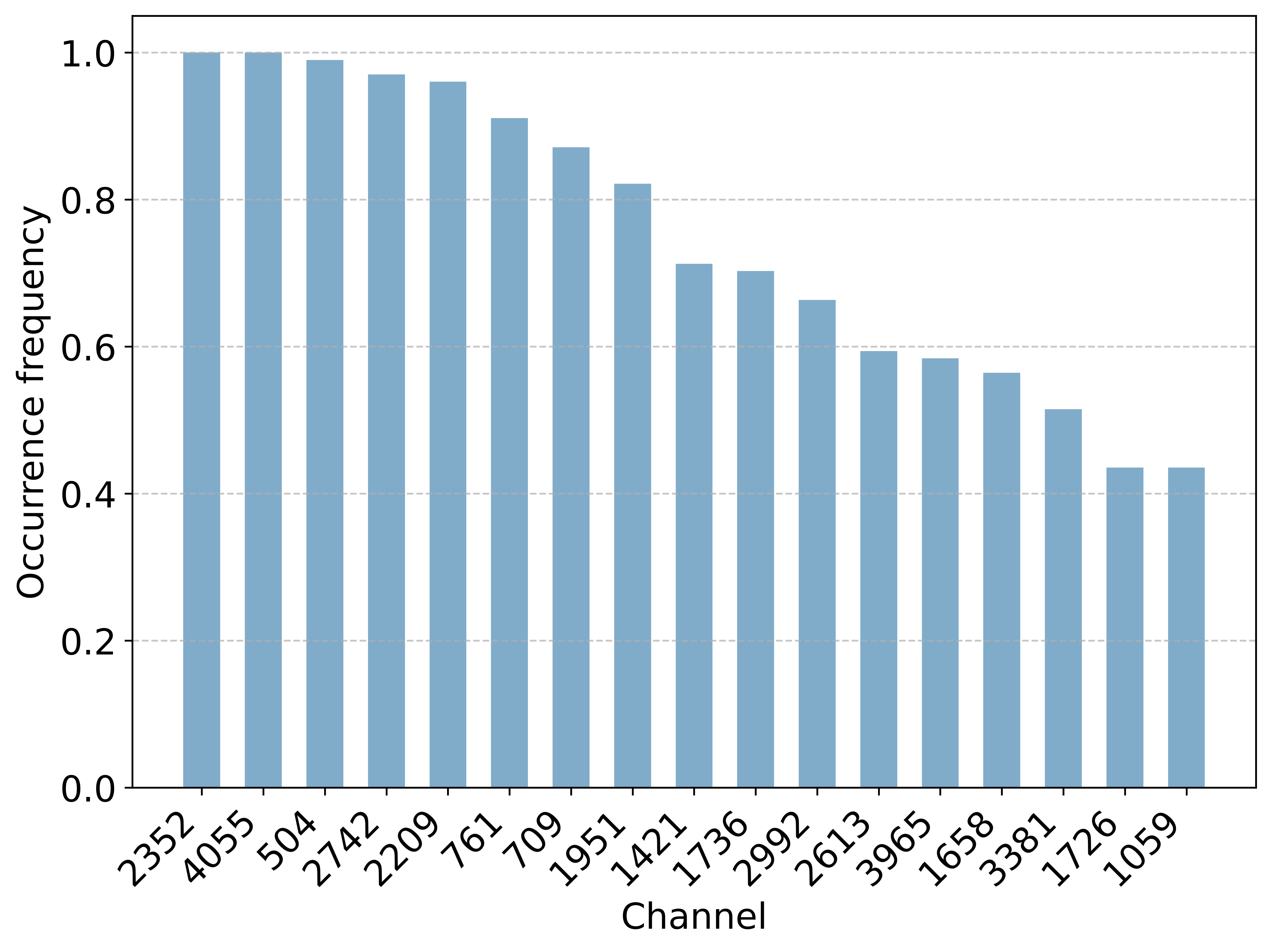}
    \caption{The occurrence frequency of pivotal channels between layer22 and layer23.}
    \label{fig:sub14}
\end{subfigure}
\smallskip % 中等大小的垂直间距（比\vspace更符合LaTeX排版习惯）
% 第二行子图
\begin{subfigure}{0.48\linewidth}
    \centering
    \includegraphics[width=\linewidth]{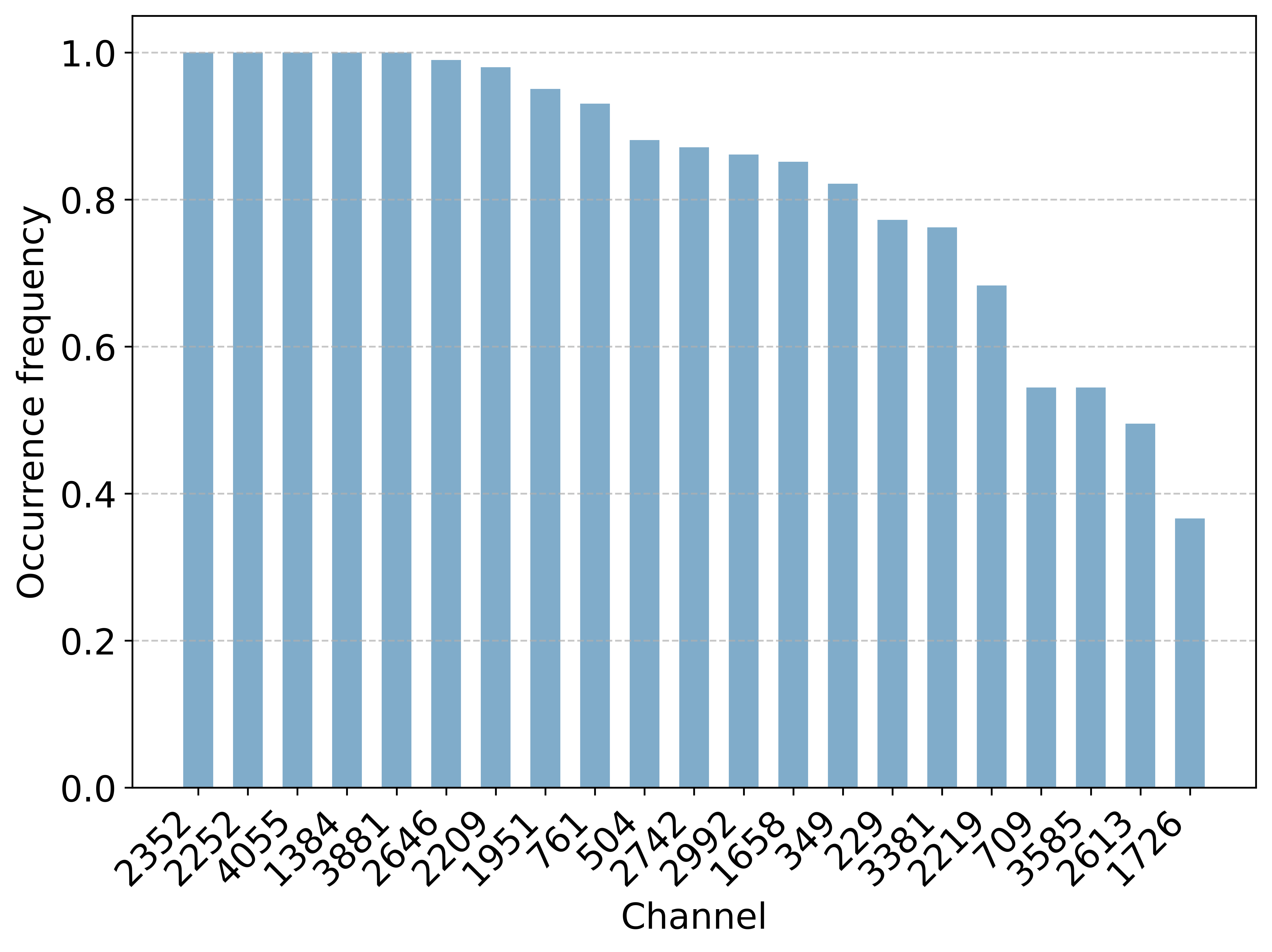}
    \caption{The occurrence frequency of pivotal channels between layer24 and layer26.}
    \label{fig:sub15}
\end{subfigure}    
\caption{The occurrence frequency of the top 0.5\% of channels by variance across layers for LLaMA3-8B. We selected 100 sentences as inputs and computed the occurrence frequencies of distinct channels based on their output hidden states.}
\label{fig5}
\end{figure}

\begin{figure}[t]
    \centering
    % 使用subfigure环境并明确设置宽度，确保每行两个子图
    \begin{subfigure}{0.48\linewidth}
      \centering
      \includegraphics[width=\linewidth]{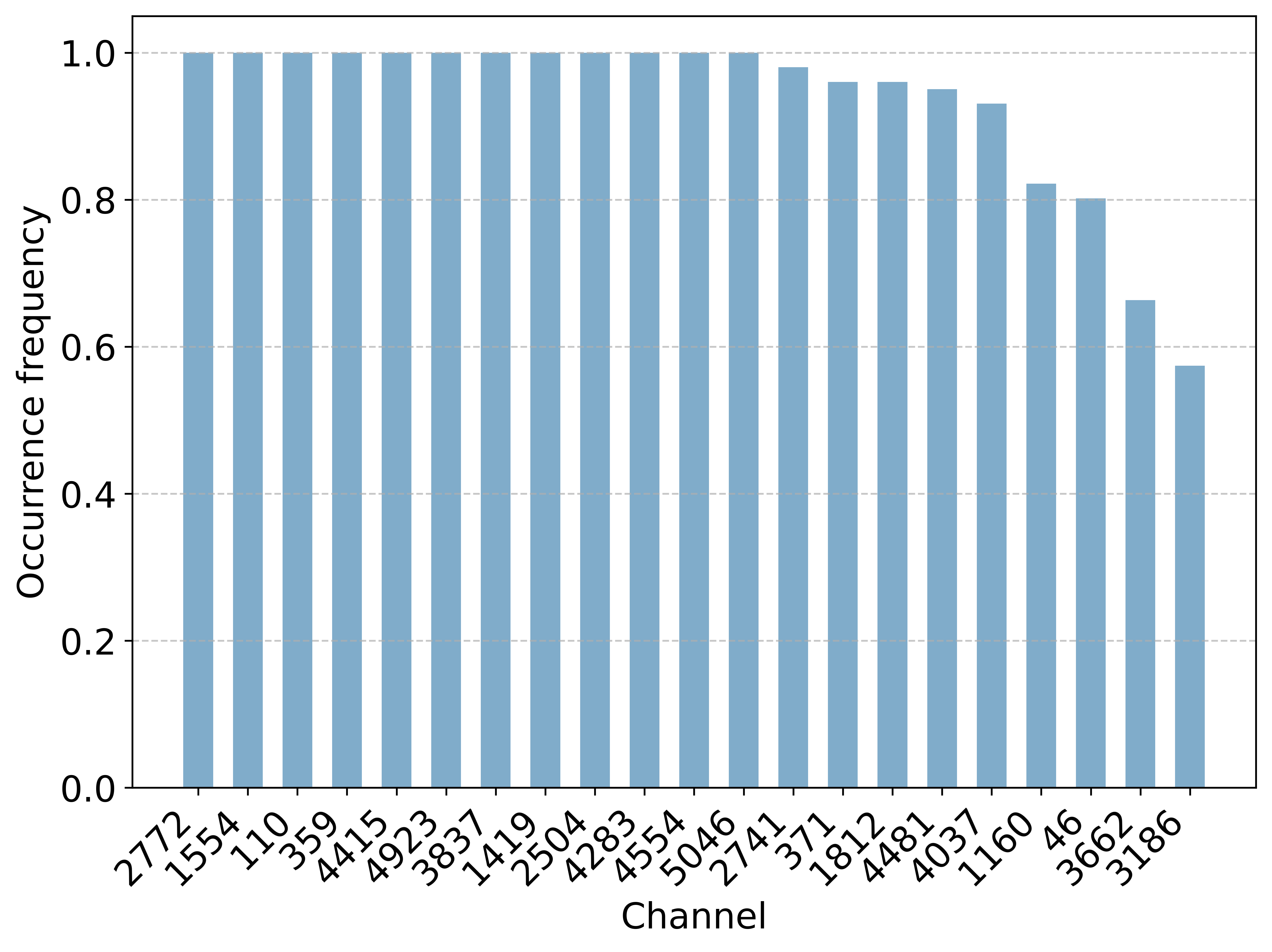}
      \caption{The occurrence frequency of pivotal channels between layer1 and layer2.}
      \label{fig:sub7}
    \end{subfigure}
    \hfill
    \begin{subfigure}{0.48\linewidth}
      \centering
      \includegraphics[width=\linewidth]{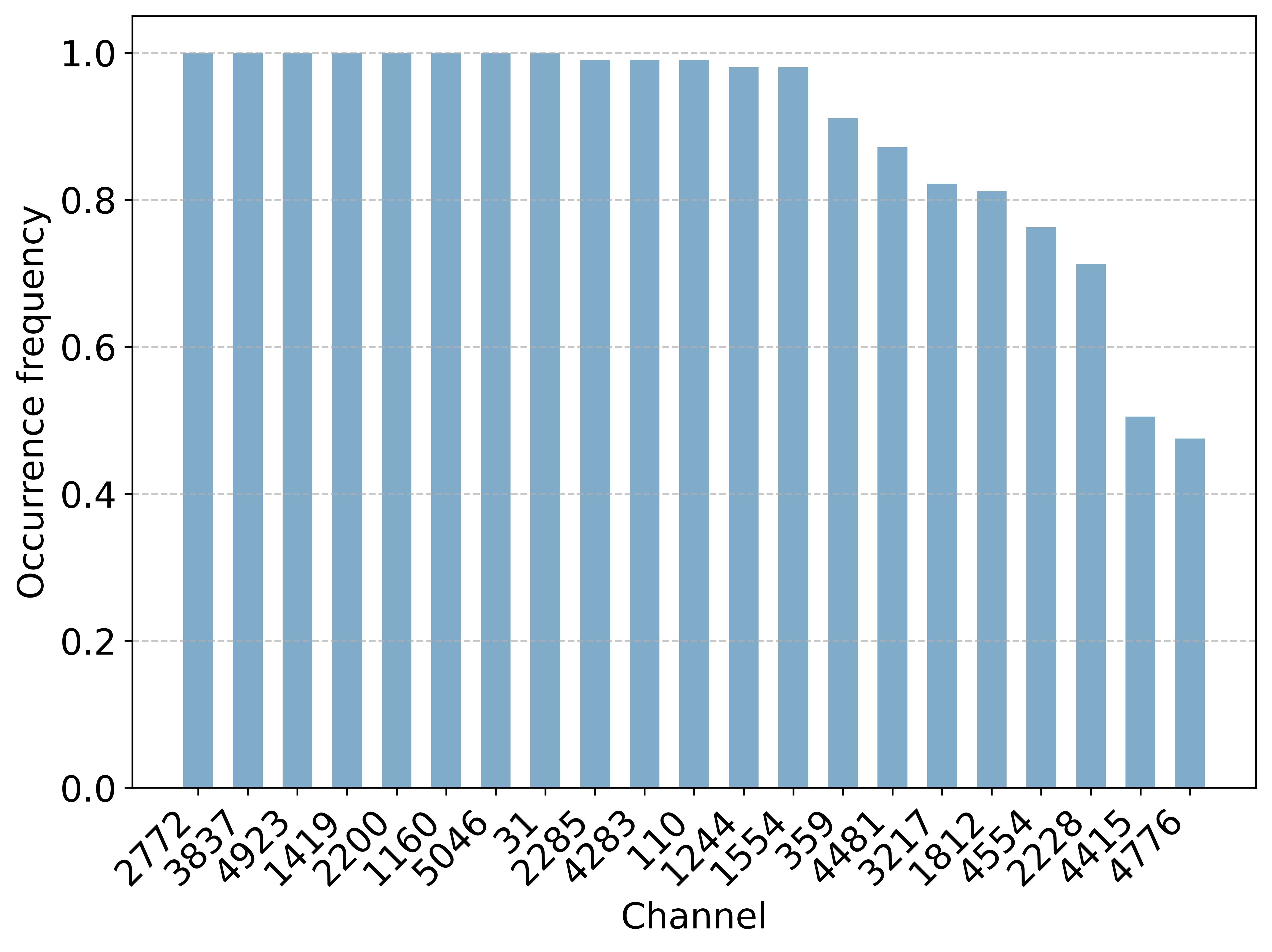}
      \caption{The occurrence frequency of pivotal channels between layer11 and layer14.}
      \label{fig:sub8}
    \end{subfigure}
    \medskip    
    \begin{subfigure}{0.48\linewidth}
      \centering
      \includegraphics[width=\linewidth]{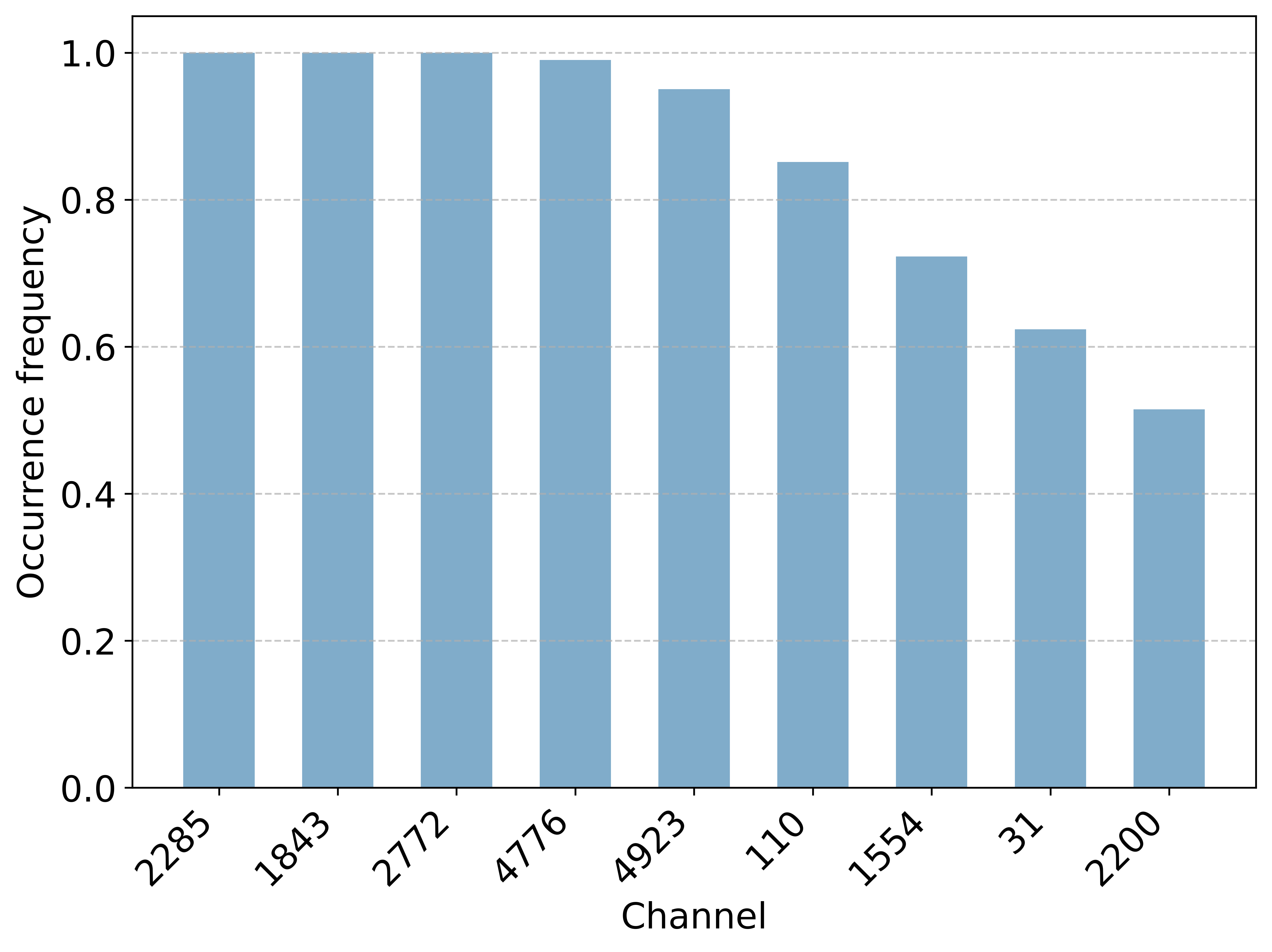}
      \caption{The occurrence frequency of pivotal channels between layer26 and layer27.}
      \label{fig:sub9}
    \end{subfigure}
    \hfill
    \begin{subfigure}{0.48\linewidth}
      \centering
      \includegraphics[width=\linewidth]{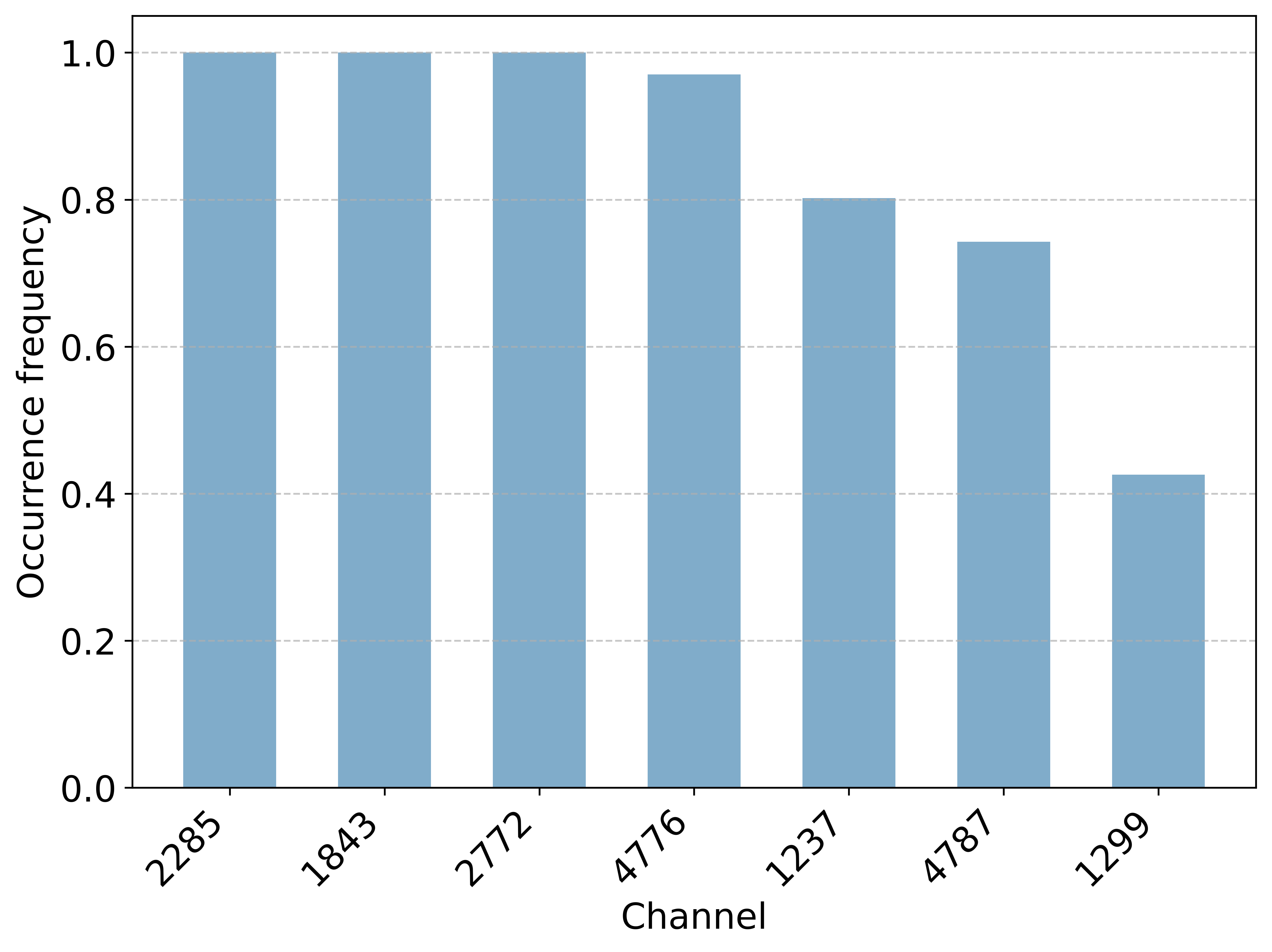}
      \caption{The occurrence frequency of pivotal channels between layer28 and layer29.}
      \label{fig:sub10}
    \end{subfigure}
    \medskip    
    \begin{subfigure}{0.48\linewidth}
      \centering
      \includegraphics[width=\linewidth]{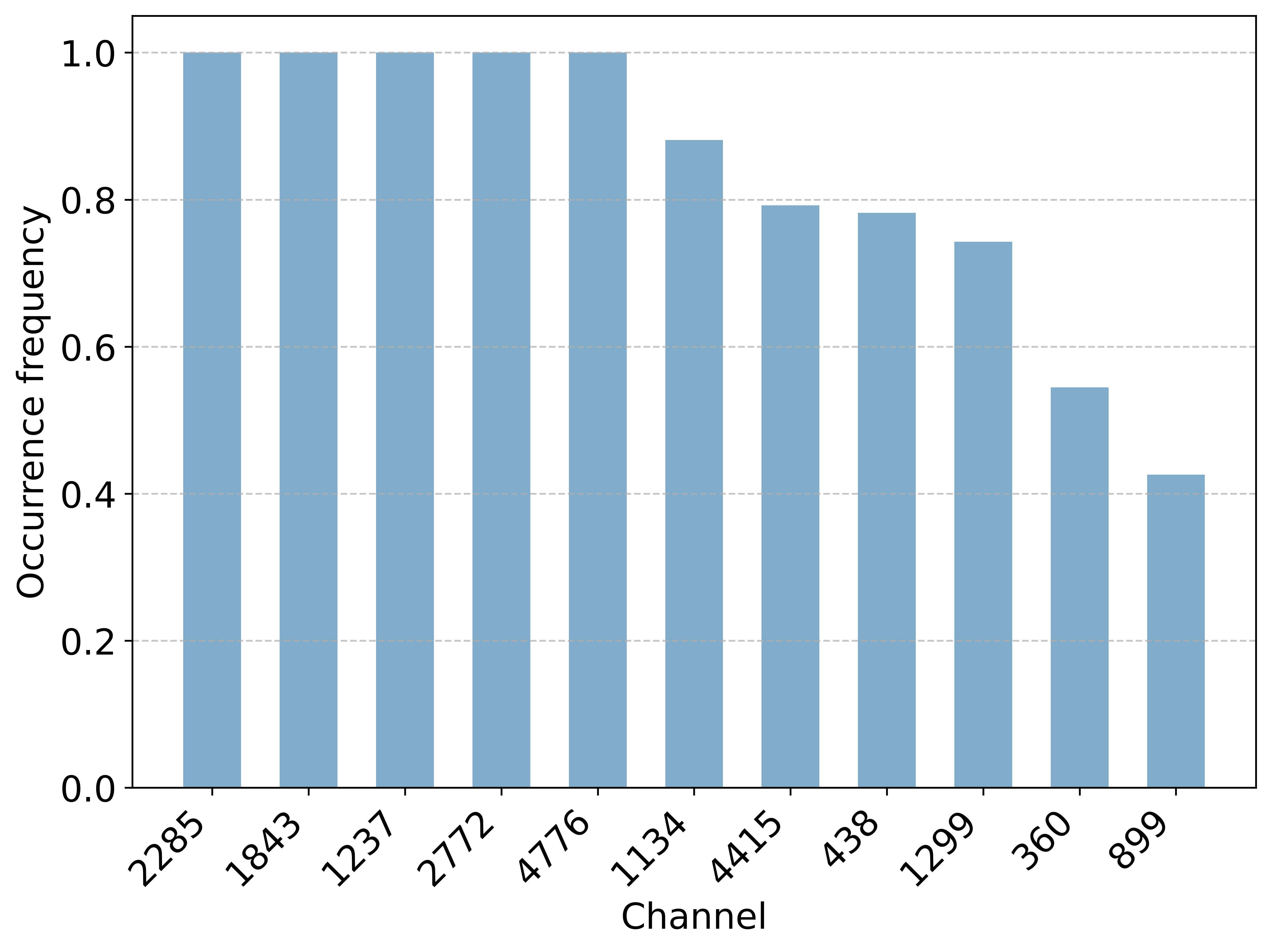}
      \caption{The occurrence frequency of pivotal channels between layer30 and layer31.}
      \label{fig:sub11}
    \end{subfigure}
    \hfill
    \begin{subfigure}{0.48\linewidth}
      \centering
      \includegraphics[width=\linewidth]{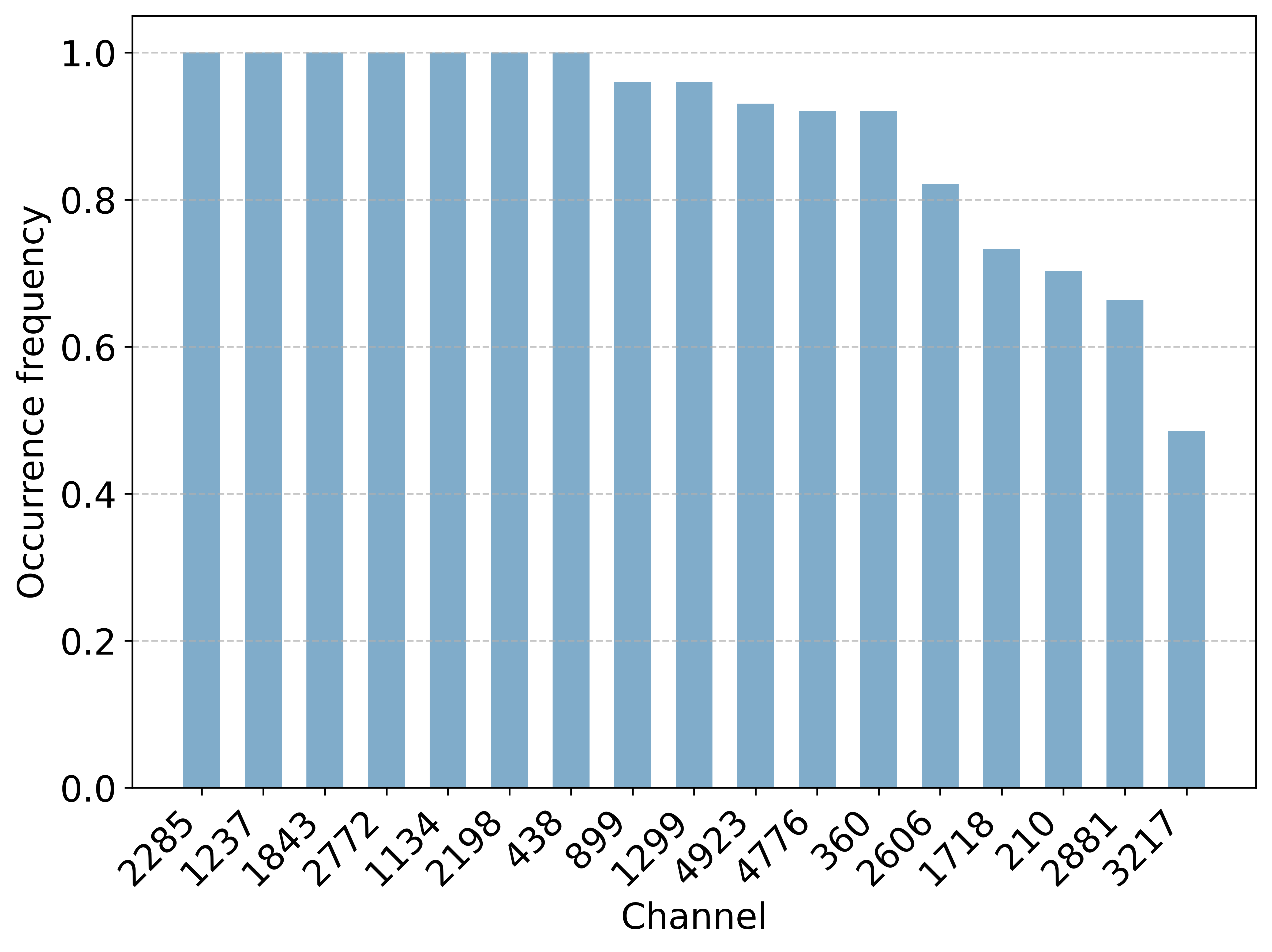}
      \caption{The occurrence frequency of pivotal channels between layer32 and layer33.}
      \label{fig:sub12}
    \end{subfigure}
    \caption{The occurrence frequency of the top 0.5\% of channels by variance across layers for LLaMA2-13B. We selected 100 sentences as inputs and computed the occurrence frequencies of distinct channels based on their output hidden states.}
    \label{fig4}
    \end{figure}

As shown in Figure~\ref{fig5} and Figure~\ref{fig4}, we selected 100 sentences as inputs from the calibration set and recorded top 0.5\% of channels by variance in hidden state outputs across different layers of the model for each input. Our experimental results demonstrate that for different LLMs and inputs, the most significantly changed channels tend to cluster in specific channels when the intermediate layer is fixed.

\newpage

\section{Supplementary Analysis on Inter-Layer Pearson Correlation Coefficients of Pivotal Channels}
\label{app2}
As shown in Table~\ref{table11}, Table~\ref{table12}, Table~\ref{table13} and Table~\ref{table14}, we calculated the Pearson correlation coefficients of the output hidden states of inter-layer pivotal channels.  It can be observed that these pivotal channels exhibit extremely high Pearson correlation coefficients, which provides sufficient theoretical justification for applying linear compensation via our PCC method.

\begin{table*}[ht]
    \centering
    \resizebox{\linewidth}{!}{
    \begin{tabular}{c c c | c c c | c c c | c c c} % 去除竖线，改为四列左对齐
        \toprule % 顶部粗线
        Layer & Channel & Pearson & Layer & Channel & Pearson & Layer & Channel & Pearson & Layer & Channel & Pearson  \\
        \midrule % 中间线
        \multirow{5}{*}{\centering 9,11} & 1512 & 0.969  & \multirow{7}{*}{\centering 13,15} & 1512 & 0.948 & \multirow{6}{*}{\centering 22,24} & 1512 & 0.961 & \multirow{11}{*}{\centering 26,27} & 1512 & 0.993 \\
        & 2158 & 0.981 & & 2789 & 0.981 & & 2883 & 0.983 & & 2883 & 0.994 \\
        & 2350 & 0.976 & & 3209 & 0.982 & & 2789 & 0.995 & & 149 & 0.965 \\
        & 2298 & 0.992 & & 1076 & 0.970 & & 2393 & 0.995 & & 3444 & 0.970 \\
        & 2235 & 0.976 & & 1404 & 0.982 & & 2622 & 0.986 & & 2639 & 0.964 \\
        & & & & 3431 & 0.984 & & 3431 & 0.989 & & 2617 & 0.958 \\
        & & & & 363 & 0.955 & & & & & 470 & 0.971 \\
        & & & & & & & & & & 2924 & 0.965 \\
        & & & & & & & & & & 3061 & 0.965 \\
        & & & & & & & & & & 2281 & 0.985 \\
        & & & & & & & & & & 2084 & 0.963 \\

        \bottomrule % 底部粗线
\end{tabular}
}
\caption{The Pearson correlation coefficient of pivotal channels in LLaMA2-7B.}
\label{table11}
\end{table*}

\begin{table*}[ht]
    \centering
    \begin{tabular}{c c c | c c c } % 去除竖线，改为四列左对齐
        \toprule % 顶部粗线
        Layer & Channel & Pearson & Layer & Channel & Pearson  \\
        \midrule % 中间线
        \multirow{9}{*}{\centering 1,2} & 2772 & 0.885  & \multirow{14}{*}{\centering 11,14} & 2772 & 0.829 \\
        & 1554 & 0.881 & & 3837 & 0.977 \\
        & 359 & 0.935 & & 4923 & 0.978 \\
        & 4923 & 0.968 & & 1419 & 0.965 \\
        & 4554 & 0.972 & & 2200 & 0.962 \\
        & 5046 & 0.969 & & 1160 & 0.977 \\
        & 2741 & 0.917 & & 2285 & 0.894 \\
        & 1812 & 0.919 & & 1554 & 0.959 \\
        & 4037 & 0.962 & & 359 & 0.977 \\
        & & & & 4481 & 0.968 \\
        & & & & 3217 & 0.867  \\
        & & & & 1812 & 0.911  \\
        & & & & 4776 & 0.959   \\
        & & & & 2741 & 0.980   \\
        \bottomrule % 底部粗线
\end{tabular}
\caption{The Pearson correlation coefficient of pivotal channels in LLaMA2-13B.}
\label{table12}
\end{table*}

\begin{table*}[ht]
    \centering
    \resizebox{\linewidth}{!}{
    \begin{tabular}{c c c | c c c | c c c | c c c} % 去除竖线，改为四列左对齐
        \toprule % 顶部粗线
        Layer & Channel & Pearson & Layer & Channel & Pearson & Layer & Channel & Pearson & Layer & Channel & Pearson \\
        \midrule % 中间线
        \multirow{7}{*}{\centering 26,27} & 1843 & 0.993  & \multirow{6}{*}{\centering 28,29} & 2285 & 0.997  & \multirow{7}{*}{\centering 30,31} & 1843 & 0.996 & \multirow{9}{*}{\centering 32,33} & 1237 & 0.951 \\
        & 2772 & 0.995 & & 1843 & 0.996 & & 1237 & 0.968 & & 1843 & 0.993 \\
        & 110 & 0.998  & & 1237 & 0.967 & & 1134 & 0.962 & & 1134 & 0.964 \\
        & 31 & 0.998  & & 1134 & 0.965 & & 4415 & 0.995 & & 2198 & 0.970 \\
        & 2200 & 0.997  & & 438 & 0.987 & & 438 & 0.988 & & 438 & 0.990 \\
        & 1718 & 0.978  & & 1718 & 0.983 & & 1554 & 0.998 & & 2606 & 0.977 \\
        & 1237 & 0.967  & & & & & 2200 & 0.996 & & 210 & 0.985 \\
        & & & & & & & & & & 2881 & 0.979 \\
        & & & & & & & & & & 110 & 0.997  \\
        \bottomrule % 底部粗线
\end{tabular}
}
\caption{The Pearson correlation coefficient of pivotal channels in LLaMA2-13B.}
\label{table13}
\end{table*}

\begin{table*}[h!]
    \centering
    \begin{tabular}{c c c | c c c | c c c} % 去除竖线，改为四列左对齐
        \toprule % 顶部粗线
        Layer & Channel & Pearson & Layer & Channel & Pearson & Layer & Channel & Pearson \\
        \midrule % 中间线
        \multirow{18}{*}{\centering 8,12} & 2932 & 0.891  & \multirow{12}{*}{\centering 22,23} & 2352 & 0.979  & \multirow{8}{*}{\centering 24,26} & 2352 & 0.979  \\
        & 1189 & 0.922 & & 4055 & 0.965 & & 2252 & 0.936  \\
        & 1298 & 0.838 & & 2209 & 0.983 & & 1384 & 0.999  \\
        & 291 & 0.891 & & 709 & 0.996 & & 1658 & 0.996  \\
        & 3266 & 0.910 & & 1951 & 0.990 & & 349 & 0.947  \\
        & 1162 & 0.868 & & 1736 & 0.966 & & 229 & 0.951  \\
        & 133 & 0.910  & & 2613 & 0.992 & & 3585 & 0.996  \\
        & 2742 & 0.685  & & 3965 & 0.966 & & 2613 & 0.992 \\
        & 1039 & 0.895  & & 1658 & 0.996 & & & \\
        & 3231 & 0.912  & & 1059 & 0.994 & & & \\
        & 2720 & 0.854  & & 357 & 0.973 & & & \\
        & 761 & 0.793  & & 2252 & 0.936 & & & \\
        & 1815 & 0.918  & & & & & & \\
        & 2977 & 0.873  & & & & & & \\
        & 3516 & 0.871  & & & & & & \\
        & 2692 & 0.885  & & & & & & \\
        & 709 & 0.916  & & & & & & \\
        & 3881 & 0.932  & & & & & & \\
        \bottomrule % 底部粗线
\end{tabular}
\caption{Pearson correlation coefficient of pivotal channels in LLaMA3-8B.}
\label{table14}
\end{table*}




% 	\bibitem{keshet2016prediction}
% 	Keshet, Renato, Alina Maor, and George Kour.
% 	\newblock Prediction-Based, Prioritized Market-Share Insight Extraction.
% 	\newblock In {\em Advanced Data Mining and Applications (ADMA), 2016 12th International 
%                       Conference of}, pages 81--94,2016.

% \end{thebibliography}

\end{document}